%% file: ICLR-ML-IRL.tex
\documentclass{article} 
\pdfoutput=1
\usepackage{iclr2020_conference,times}
\usepackage{amsmath}
\input{math_commands.tex}

\usepackage{epsfig}
\usepackage{url}
\usepackage{tabularx,booktabs,multicol}
\newcolumntype{C}{>{\centering\arraybackslash}X} 
\usepackage{graphicx}
\usepackage{caption}
\usepackage{subcaption}
\usepackage{multirow}
\usepackage{adjustbox}
\usepackage{float}
\usepackage{makecell}
\usepackage{amssymb}
\usepackage{verbatim}

\usepackage[colorinlistoftodos]{todonotes}
\usepackage[colorlinks=true, linkcolor=blue,citecolor=green]{hyperref}

\title{\textit{Egoshots}, an ego-vision life-logging dataset and semantic fidelity metric to evaluate \\diversity in image captioning models
}

\author{Pranav Agarwal \hspace{75pt} Alejandro Betancourt \hspace{35pt} Vana Panagiotou\\
INRIA RITS, Paris, France\hspace{41pt}LANDING AI,\hspace{73pt}Signal Ocean,\\
\texttt{pranav2109@hotmail.com}\hspace{16pt} Colombia\hspace{92pt} Greece\\
\And
\hspace{145pt}
Natalia D\'iaz-Rodr\'iguez\thanks{Work partially done during D\'iaz-Rodr\'iguez and Panagiotou's internship at Philips Research, and Betancourt's stay at Eindhoven University of Technology, Netherlands.} \\
\hspace{55pt}U2IS, ENSTA, Institut Polytechnique Paris and INRIA Flowers, France\\
\hspace{120pt}\texttt{natalia.diaz@ensta-paris.fr}\\
}

\iclrfinalcopy 
\begin{document}

\maketitle

\begin{abstract}
Image captioning models have been able to generate grammatically correct and human understandable sentences. However most of the captions convey limited information as the model used is trained on datasets that do not caption all possible objects existing in everyday life. Due to this lack of prior information most of the captions are biased to only a few objects present in the scene, hence limiting their usage in daily life. In this paper, we attempt to show the biased nature of the currently existing image captioning models and present a new image captioning dataset, \textit{Egoshots}, consisting of 978 real life images with no captions. We further exploit the state of the art pre-trained image captioning and object recognition networks to annotate our images and show the limitations of existing works. Furthermore, in order to evaluate the quality of the generated captions, we propose a new image captioning metric, object based \textit{Semantic Fidelity} (SF). Existing image captioning metrics can evaluate a caption only in the presence of their corresponding annotations; however, SF allows evaluating captions generated for images without annotations, making it highly useful for real life generated captions.

\end{abstract}

\section{Introduction}
Humans have a great ability to comprehend any new scene captured by their eyes. With the recent advancement of deep learning, the same ability has been shared with machines. This ability to describe any image in the form of a sentence is also well known as \textit{image captioning} and has been at the forefront of research for both computer vision and natural language processing \citep{Vinyals2014ShowAT,Karpathy2014DeepVA,Venugopalan2016CaptioningIW,Selvaraju2019TakingAH,Vedantam2017ContextAwareCF}. The combination of Convolutional Neural Networks (CNN) and Recurrent Neural Network (RNN) has played a major role in achieving close to human like performance, where the former maps the high dimensional image to efficient low dimensional features, and the latter use this low dimensional features to generate captions. However, most of the image captioning models have mostly been trained on MSCOCO \citep{Lin2014MicrosoftCC} or Pascal-VOC \citep{pascal-voc-2012}, which consists of 80 and 20 object classes respectively. All the images are captioned taking into consideration only these classes. Thus, even though current models have been successful in generating grammatically correct sentences, they still give a poor interpretation of a scene because of the lack of knowledge about various other kinds of objects present in the world, along with those seen in the dataset. Egoshots dataset\footnote{Egoshots dataset available at \url{https://github.com/NataliaDiaz/Egoshots}} has a wide variety of images ranging from both indoor to outdoor scenes and takes into consideration diverse situations encountered in real life which are hardly found in the MSCOCO or Pascal-VOC dataset. Hence our dataset can act as a benchmark to evaluate the performance and robustness of image captioning models on real-life images. 

Image captioning has a wide range of applications from supporting a visually impaired person to the recommendation system. Guiding a visually challenged person \citep{Gurari2018VizWizGC} is a highly sensitive task and a small error can lead to catastrophic accidents. A recent work \citep{Weiss19} has shown the ability to successfully complete this task in a simulated environment using reinforcement learning. Image captioning models can also play a major role in further improving these systems. However, in order to make these systems more reliable, the network trained should be able to predict considerably more descriptive captions taking into consideration all the objects present in the scene. 
In this work, along with releasing the captioned Egoshots dataset, 
we show how pre-trained networks perform on different animated pictures or paintings where it is easy for humans to interpret the scene, but a neural network completely fails, and propose a metric of semantic fidelity to the actual scene being rendered. The captions predicted for artistic images conclude that the current image captioning networks lack robustness and are biased towards a few annotated object classes and the kind of controlled scenes present only in training datasets, hence they are not reliable and cannot be deployed for real-life applications without significant finetuning.


Current metrics are limited to evaluate the quality only in the presence of labels rendering them useless for captions generated for real-life scenes. We aim at tackling both these problems by proposing the Egoshots dataset and SF metric. Egoshots consists of 978 real-life ego-vision images captioned using state of the art image captioning models, and aims at evaluating the robustness, diversity, and sensitivity of these models, as well as providing an on-the-wild life-logging dataset that can aid the task of evaluating real life settings. Images were randomly taken by the \textit{Autographer} camera worn by 2 female computer science interns (aged 25 and 29) for 1 month each in Eindhoven (Netherlands), doing regular activities (biking, office working, socializing...) during May-Jul 2015. We further propose a new metric SF so that captions generated by pre-trained models can be evaluated both on their relevance to the image being captioned, and the number of different objects the caption takes into consideration.  


\begin{figure}[t]
\begin{center}
\includegraphics[width=1.0\linewidth]{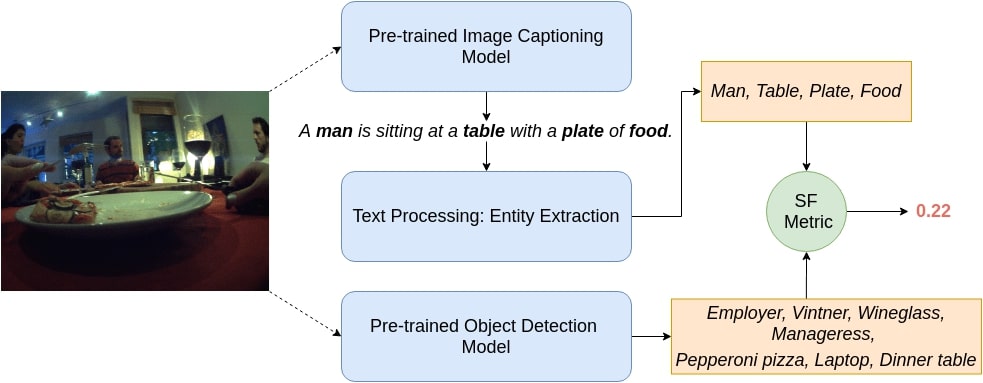}
\caption{The overview of our approach for captioning the Egoshots dataset and further evaluating the caption using our proposed Semantic Fidelity (SF) metric. The image to be captioned is pre-processed and passed through a pre-trained image captioning model. The caption generated is processed to output only nouns. The same image is also input to an Object Detector (OD), which detects all the object classes in the given image. Both the noun and object classes detected are then used to compute the SF for a generated caption.}
\label{fig:0}
\end{center}
\end{figure}

\section{Annotation Pipeline and Semantic Fidelity Metric}


\textbf{Annotation Pipeline}: Transfer learning has proven to be an effective approach to perform a known task in a new environment efficiently \citep{Tan2018ASO}. The idea is to 
use a pre-trained model on a new dataset, in order to prevent learning features from scratch, which requires excessive computational resources and time. To caption Egoshots dataset we follow this methodology and use pre-trained weights of state of the art image captioning models without any finetuning.  We restrict our work to three models, namely \textit{Show Attend And Tell} (SAT), \textit{nocaps: novel object captioning at scale} (NOC), and \textit{Decoupled Novel Object Captioner} (DNOC), as they were able to achieve the best results for the real-life Egoshots dataset. Since pre-trained neural networks are strictly constrained to the size of input images on which they were trained, we use the exact pre-processing for images used in each of the models. Once processed, the images are mapped to their corresponding captions using the learnt models. Among the captions predicted, the one having maximum SF is used as the final caption for the dataset. Section \ref{sec:ba} further describes in detail each of the image captioning models. 

\textbf{Object based Semantic Fidelity Metric}: To render caption-less datasets such as Egoshots useful, the aim is to map each image to the 
richest caption, i.e., the one covering 
as many objects as relevant in the scene, i.e., a caption as descriptive and detailed as possible. With current image captioning metrics, this task renders challenging, 
since all existing metrics evaluate the quality of generated captions using labeled captions from a dataset collected in 
well controlled and clean conditions (very different for real life first person vision images). In order to counterbalance this assumption not present in Egoshots, we propose a new image captioning metric called \textit{Semantic Fidelity}. SF takes into consideration two elements: 1) the semantic closeness of the generated caption to the objects detected in the image, and 2) the object diversity with respect to the amount of object instances detected. 
Assuming a state-of-the-art quasi perfect object detector, by taking into account the semantic closeness among these two sets of (captioned and detected) entities (i.e. objects), we penalize the model when it predicts a caption containing objects completely different from the objects present in the scene. 


For each caption generated by a captioning model, we eliminate all words except nouns. This is done as a simplification, assuming nouns convey the largest information on the number of different objects present \citep{Wang2018ObjectCB}. Let $C = \{c_1,....,c_m\}$ be a list of captions generated by the pretrained networks, and $W = \{w_{1},....,w_k\}$ the generated words for a given caption $c_i$ (further processed to keep only nouns for the given caption $i$ $N_i = \{n_{1},....,n_{z}\}$. In order to predict all objects present in the image we use a state of the art object detector (OD)\footnote{We tested several different ODs and settled with YOLO9000 (or Y9) in this particular showcase of SF metric because of its ability to detect the largest (9000) number of classes, while other detectors are restricted to either 80 or 20 classes only, as shown in Fig. \ref{fig:sub2}.}. 
The objects predicted by the OD are represented as $O_{OD} = \{o_{1},....,o_{y}\}$. Thus, for every image we obtain a list of noun words from its predicted caption, and a list of objects detected by the OD. A similarity metric among these word sets\footnote{Such as the cosine similarity (as we use here) of the mean of the embeddings of the words in each of these two sets.} 
is calculated. The semantic similarity among these two word sets takes into account their semantic closeness using word embeddings. Recent works \citep{Mikolov2013DistributedRO,Conneau2017WordTW} show the ability of word embeddings that is transforming a word into its vectored form efficiently capture the semantic closeness of two given words. The SF metric uses this approach to calculate such semantic similarity between the noun words and 
objects in an image, for each caption, described as $S = \{s_{1},....,s_{m}\}$.
The cosine similarity as such does not take into account the diverse nature of the predicted captions in terms of the number of different objects present, since the final embedding calculated for each of the list averages out multiple objects of the same class as a single entity. Hence, to further penalise the quality of captions for its object count we compute the ratio of the number of noun words in a given caption to the number of objects predicted. The SF metric score thus measures the quality of a caption, as a proxy 
of the diversity of knowledge the network has regarding both the presence 
and diversity of different objects present in the image: 
\begin{equation}
\label{eqn:3}     
    SF_i = {s_i}\ldotp{\frac{\#N}{\#O}}
\end{equation}
where, for image $i$, $s_i$ is the semantic similarity among noun words in its predicted caption $c_i$ and object nouns detected by the OD, $\#O$ is the cardinal of $O_{OD}$, 
and $\#N$ the number of nouns (representing objects in $N_i$) present in $c_i$. SF ranges in [0, 1]: captions having an SF closer to 1 convey more information and are semantically closer to the scene being captioned, in terms of the objects involved in the caption.

For a predicted caption, information about an 
average of 2-3 objects is generally conveyed (as shown in Fig. \ref{fig:sub1}), while for the OD we assume an ideal condition in which the OD acts as an oracle and predicts all different objects in the given image, so that $\#O \geq \#N$ (Assumption 1) for all images. 
Thus, the larger the number of noun entities in the caption, the more the ratio will approach 1, and 
the closer SF will be to 1 for the best captions. 
This approach to compute SF will work only 
assuming robust object detectors 
satisfying enough scene annotation granularity. For the SF values to be 
reliable, 
the OD needs to detect correct object classes as present in the image. With completely different objects detected with respect to those in the image, the similarity metric for image $i$, $s_i$, will be inaccurate, and therefore, 
SF remains unreliable. In order for SF to be applicable, we will also assume Assumption 2: $\#O \neq 0$ (i.e., the OD can at least detect one object in the image). 
In scenarios when the object detector predicts less objects than nouns in the sentence (Assumption 1 broken), we skip the object diversity ratio and use $SF = s_i$ instead.

\section{Discussion and Conclusion}\label{sec:ab}

Table \ref{ref:label3} compares the SF performance on different image captioning methods using various object detectors (OD). To make SF reliable, we assume an OD to predict all the relevant different classes given in each image. 
Except Y9, all other ODs are trained only on either 80 or 20 classes, leading to larger SF values for weak detectors, as they find it difficult to penalize an inefficient caption given their lack of knowledge about various different object classes. Also ODs trained on 20 classes of VOC tend to show larger SF than those trained on 80 classes of COCO because of their inability to detect new classes not seen in the training dataset\footnote{Or classes not always visually \textit{observable}. An example of such words is, e.g. the class \textit{entrepreneur} from YOLO9000.}. Better OD models will make SF more reliable, as is also reflected for the captions generated in Table \ref{label1}. Table \ref{label1} compares the performance of each of the pre-trained image captioning models on the SF metric. OD Y9 for image \textcolor{blue}{3} in Table \ref{label1} predicted a single incorrect class 
leading to inaccurate SF value due to using a similarity metric among wrong terms. 
However, 
for most of the other images, Y9 is able to predict 
the correct objects; therefore SF is able to penalize the captions which do not take into consideration all the objects present in a given scene. Hence, a well generalized and robust object detection model plays the most important role if the evaluation of captions is performed using SF. 


Most images are captioned using 8-10 words only. Even if longer sentences 
may not always be preferred, 
critical applications, e.g., supporting the visually impaired or autonomous driving, where the agent does not have any prior information, the 
richer the interpretation of a scene, the more reliable the system will be. 

We presented the \textit{Egoshots} dataset, consisting of 978 real life first person vision images of everyday activities. Although the number of images in the dataset is far less than those in MSCOCO or Pascal VOC, our aim, along with presenting the dataset, is to analyze the performance of the pre-existing image captioning models and their reliability. To do so we propose the first image captioning metric (SF) that allows to evaluate captions from unlabelled images. Since previously existing metrics are limited to caption-labelled datasets, there is no way to analyze the quality of captions of real life images without costly annotated labels. 
We show on pre-trained models that, despite being able to successfully generate grammatically correct sentences, their captions are often misaligned with the objects present in the scene or hallucinate objects \citep{Rohrbach18}. This is due to the presence of training bias towards objects imposed by the datasets these networks were trained with:  
they are unable to convey complete or rich information about the scene. 
The aim of Egoshots dataset and SF metric is thus to facilitate assessment, diversity and deployability of image captioning deep models. 

Despite the major improvements in recent image captioning models, 
we show that 
there is room to achieve more semantically faithful and relevant caption generation systems; e.g. models could be affected by a significant amount of blurring, making object detection in the wild a lot harder. Thus, there are plenty of room for future work on OD and IC models on real life-logging situations, not only for assisted technology or telepresence, but for any autonomous system. Since the SF score's aim is to measure the semantic quality of a caption, and it cannot evaluate the grammatical correctness of a sentence, future work should better assess both detector and captioning models quality 
to better map the desirable properties of such models at a finer grained domain-specific resolution. Such studies should include larger scale assessments with a positive control group (i.e., human annotations of real images on the wild).


\bibliography{refs}
\bibliographystyle{iclr2020_conference}
\newpage
\appendix
\section{Appendix}

\subsection{Related Work}

\subsubsection{Image Captioning}

The problem of scene understanding and image captioning has been vastly explored in the literature \citep{Zhang2020InteractionGF,Nagarajan2020EGOTOPOEA,Tsutsui2019ActiveOM,Liang2020VisualSemanticGA,Kenigsfield2019LeveragingAT}. Most of the work follows a sequence learning approach using an encoder and decoder \citep{Vinyals2014ShowAT,Olivastri2019EndtoEndVC,Donahue2015LongtermRC,Fan2016DeepDiaryAC}. The encoder consists of several stacked CNNs to achieve efficient latent representations of images, while the decoder has recurrent layers to map these latent vectors to a caption. In order to make the captions more diverse, the attention mechanism integrated into the encoder-decoder framework \citep{Zhang2020InteractionGF,You2016ImageCW,Li2017ImageCW,Lu2018NeuralBT,Anderson2017BottomUpAT,Selvaraju2019TakingAH,Lu2016KnowingWT}, which takes into consideration long-range dependencies. Attention gives importance to different spatial regions of the image by weighing them differently. The decoder then generates each word by taking into consideration the relative importance of each spatial region. Most approaches follow this methodology and train the models  using datasets such as MS-COCO \citep{Lin2014MicrosoftCC}, PASCAL-VOC \citep{pascal-voc-2012} and Flickr 30k \citep{Young2014FromID} to name a few. These datasets have millions of images with human labelled captions for a predefined number of object categories. COCO dataset has captions for 80 different object classes, while PASCAL-VOC has 20 different classes. Some of the recent works \citep{Venugopalan2016CaptioningIW,Wu2018DecoupledNO,Demirel2019ImageCW} have tried to break this limitation by integrating different object classes while training their image captioning model. They predict a placeholder in the generated caption for the object classes not present in the MSCOCO dataset and further replace this placeholder with the object classes. This approach was able to generate more diverse captions, to some extent, than previous models; however, it still lacks the ability to produce captions able to take into consideration all different objects present in a scene.

There have been very few works which have used unsupervised \citep{Feng2018UnsupervisedIC} or reinforcement learning \citep{Ren2017DeepRL} for captioning images.Unsupervised image captioning model use generative adversarial networks in order to generate captions close to those created by humans. They use an external corpus of sentences in order to train the model. The reinforcement learning model uses an actor-critic approach with a reward model for mapping images to their corresponding captions. The actor predicts the confidence of predicting the next word given the image, while the critic takes in the current state and the words predicted, and adjusts the goal to produce captions close to the ground truth.

\subsubsection{Image Captioning Metrics}
In order to evaluate the quality of the generated captions most of the models use automatic image captioning metrics such as BLEU \citep{Papineni2001BleuAM}, Meteor \citep{Banerjee2005METEORAA}, Rouge \citep{Lin2004ROUGEAP} and CIDEr \citep{Vedantam2014CIDErCI}. These metrics evaluate the quality of the captions based on pre-existing labels. To the best of our knowledge there is no metric which can evaluate the quality of a captions without labels. Thus, testing the performance of an image captioning model on a real-life image not present in the dataset becomes a major limitation. On the other side, human evaluation is extensively time-consuming and not reliable, as it varies from person to person. In this work, we presented the new metric, SF, which allows to evaluate an unlabelled caption (i.e., captions for which its ground truth is not available). 

\begin{figure}[htbp!]
  \centering
  \begin{subfigure}[b]{0.7\textwidth}
    \includegraphics[width=\textwidth]{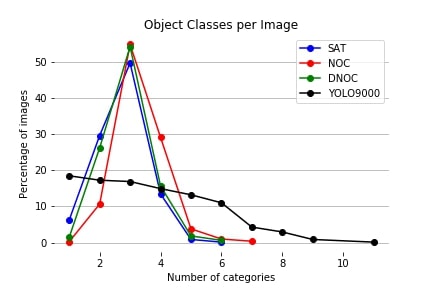}
    \caption{}
    \label{fig:sub1}
  \end{subfigure}\hspace{5mm}
  \begin{subfigure}[b]{0.7\textwidth}
    \includegraphics[width=\textwidth]{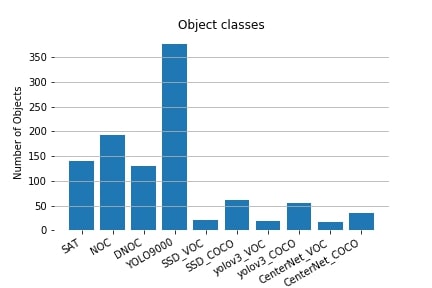}
    \caption{}
    \label{fig:sub2}
  \end{subfigure}\hspace{5mm}
  \begin{subfigure}[b]{0.7\textwidth}
    \includegraphics[width=\textwidth]{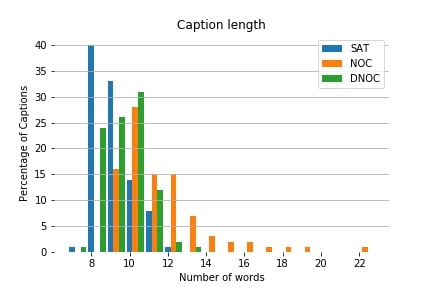}
    \caption{}
    \label{fig:sub3}
    \end{subfigure}
  \caption{a) Number of nouns per image for SAT, NOC, DNOC, and number of object categories for YOLO9000. b) Total number of different object classes present in the Egoshots dataset as predicted by each of the pre trained image captioning and object detection models. c) Caption length per image for SAT, NOC and YOLO9000.}
  \label{fig:1}
\end{figure}

\subsubsection{Image Captioning and Object Detection Models assessed}\label{sec:ba}

This section details the existing state of the art captioning models used to annotate Egoshots dataset and validate the SF metric.

\textbf{Show Attend and Tell (SAT)} \citep{Xu2015ShowAA} uses a CNN followed by an LSTM in order to caption the image. CNNs take the input image to extract efficient low dimensional features which are then further decoded by the LSTMs to generate captions in a sequence to sequence manner. In order to predict realistic captions, they integrate the attention module into LSTMs in order to focus on salient objects. Two kind of attention modules are used; namely, soft and hard attention. Even though attention helps improving captions, at the same time, attention is responsible for less descriptive captions, as with attention the network focuses on only important objects in the image while filtering away a large number of objects.  

\textbf{Novel object captioning at scale (NOC)} \citep{Agrawal2019nocapsNO} tries to tackle the problem of having fewer object classes present in the captions of the COCO dataset by incorporating the Open Image dataset \citep{Kuznetsova2018TheOI} which has 600 classes but still far less than YOLO9000. It tries to disentangle object detection and image captioning and claims to have a greater number of object classes in the generated captions in comparison to the MSCOCO but if we compare the performance on Egoshots dataset with respect to the total number of different object classes used for captioning it still lags behind YOLO9000 as shown in Fig. \ref{fig:sub1}. 

\textbf{Decoupled Novel Object Captioner (DNOC)} \citep{Wu2018DecoupledNO} follows a two-step process for generating sentences. They start with predicting captions with placeholders for every novel object not seen previously in the dataset. In the second step, they use an object memory to replace the placeholder with the correct object word based on the visual features. Pre-trained object detection is used to predict novel objects. DNOC also uses an encoder-decoder architecture with a slight variation in the decoder part. The encoder is a network pre-trained on ImageNet to extract the low dimensional features. The decoder uses LSTMs and the output from the encoder to generate word by word a sentence. They predict novel objects in the captions not seen in the dataset but still, their captions are not descriptive with respect to the number of object classes per image far less than NOC and YOLO9000. 

\textbf{YOLO9000} (Y9 for short) \citep{Redmon2016YOLO9000BF}. We use a pre-trained YOLO9000 object detector model in the Egoshots dataset as state of the art OD in order to evaluate the diversity of generated captions. YOLO9000 achieved state of the art results for object detection, and can detect around 9000 different object classes. It integrates both object detection and image classification into a single module (trained on MSCOCO and ImageNet datasets). Since ImageNet has a greater number of object classes than MSCOCO, it detects more objects than those present in MSCOCO dataset. Despite being far from being 100\% accurate, it is used as gold standard OD in our experiments to illustrate the use of SF. However, any state of the art object detector can be used/adapted to each domain specific problem, as state of the art OD. 

In Fig. \ref{fig:sub1} we show that 
most of the captions generated convey information about max. 2-4 objects for a given image. At the same time, Y9 is able to detect 5-7 objects for most of the images, and is also able to predict a maximum of 12 classes for few images. In addition to this, Fig. \ref{fig:sub2} shows that for the Egoshots dataset, Y9 can detect 370 unique objects, while the captions predicted use approximately half the number of objects.  The inability of captions to take into consideration all the objects present in the image is also shown in Fig. \ref{fig:sub3}.

Baselines implemented are available online\footnote{\url{https://github.com/Pranav21091996/Semantic_Fidelity-and-Egoshots}}.
\begin{table}[]
\begin{tabular}{lllllllll}
\multicolumn{9}{l}{} \\ \Xhline{2\arrayrulewidth}
\multicolumn{2}{l}{\textbf{Image Captioning Method}} & \textbf{S}-V & \textbf{S}-Co & \textbf{Y3}-V & \textbf{Y3}-Co & \textbf{C}-V & \textbf{C}-Co & \textbf{Y9} \\ \hline
\multicolumn{2}{l}{\textit{Show Attend And Tell}} & 0.35 & 0.34 & 0.34 & 0.33 & 0.30 & 0.36 & \textbf{0.28} \\
\multicolumn{2}{l}{\textit{Novel Object Captioning at Scale}} & 0.40 & 0.39 & 0.39 & 0.37 & 0.34 & 0.40 & \textbf{0.33} \\
\multicolumn{2}{l}{\textit{Decoupled Novel Object Captioner}} & 0.41 & 0.41 & 0.40 & 0.39 & 0.35 & 0.44 & \textbf{0.32} \\ \Xhline{2\arrayrulewidth}
\end{tabular}
\caption{Mean Semantic Fidelity of different image captioning models using various object detectors: S: \textit{SSD \citep{Liu2016SSDSS}}, Y3: \textit{YOLOv3 \citep{Redmon2018YOLOv3AI}}, C: \textit{Center Net \citep{Duan2019CenterNetKT}}, Y9: \textit{YOLO9000} trained on \textit{ImageNet} and \textit{COCO}, V: trained on \textit{VOC}, Co: trained on \textit{COCO}.}
\label{ref:label3}
\end{table}

\subsection{Validating Semantic Fidelity Metric with Human Semantic Fidelity} 

\begin{figure}[htbp!]
\begin{center}
\includegraphics[width=55mm,scale=0.5]{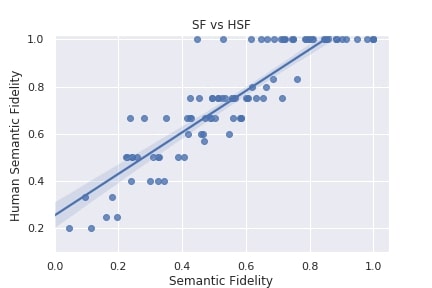}
\end{center}
\caption{Linear fitting test for SF and Human SF (HSF). Pearson correlation test for 100 MSCOCO dataset manually annotated images gives positive correlation with $\rho$ = 0.93.} 
\label{fig:6}
\end{figure}

SF metric can use different similarity  metrics. In our case we use cosine similarity in order to compute each word embedding and then calculate the semantic similarity. We use spaCy NLP toolkit \footnote{\url{https://spacy.io/}} implementation.

To further validate our SF metric, we perform a linear regression analysis by 
comparing SF scores with those SF scores provided by a human labeller (Human Semantic Fidelity, HSF). We use 100 MSCOCO dataset images. 
For SF we use MSCOCO ground truth (GT) captions, while for HSF we use human labelled image captions taken from MSCOCO 
caption, i.e.: we set HSF to be the number of objects in the caption divided by the number of real (GT) objects observed in image $i$:

\begin{equation}
\label{eqn:4}
    {HSF}_i = \frac{\#N}{\#O_{GT}}
\end{equation}

For both SF and HSF, a ground truth (GT) human annotated object detector is used to annotate the images, hence assumption 1 ($\#O \geq \#N$) is always valid.  
Pearson's correlation 
$\rho$ 
parameter (to compare the (co)relation between two independent variables) analysis showed a positive correlation among SF and HSF of 0.93, 
with a p-value of 1e-44. The coefficient of determination $R^2$ showed that approximately $76\%$ of the observed variation can be explained by the linear model of the 100 manually labelled datapoints\footnote{Available in the dataset folder in \url{https://github.com/Pranav21091996/Semantic_Fidelity-and-Egoshots}}.


\subsection{Comparing pre-trained image captioning models using Semantic Fidelity} 

The validity of the SF metric is subject to the generalization and robustness properties of the OD to new images (i.e., the higher the F-measure of the OD, the more reliable the SF will be).

The following examples illustrate the use of the Semantic Fidelity proposed metric on the Egoshots dataset, evaluating different image captioning models.

\begin{table}[H]
\textbf{1)} \adjustimage{width=3cm,valign=c}{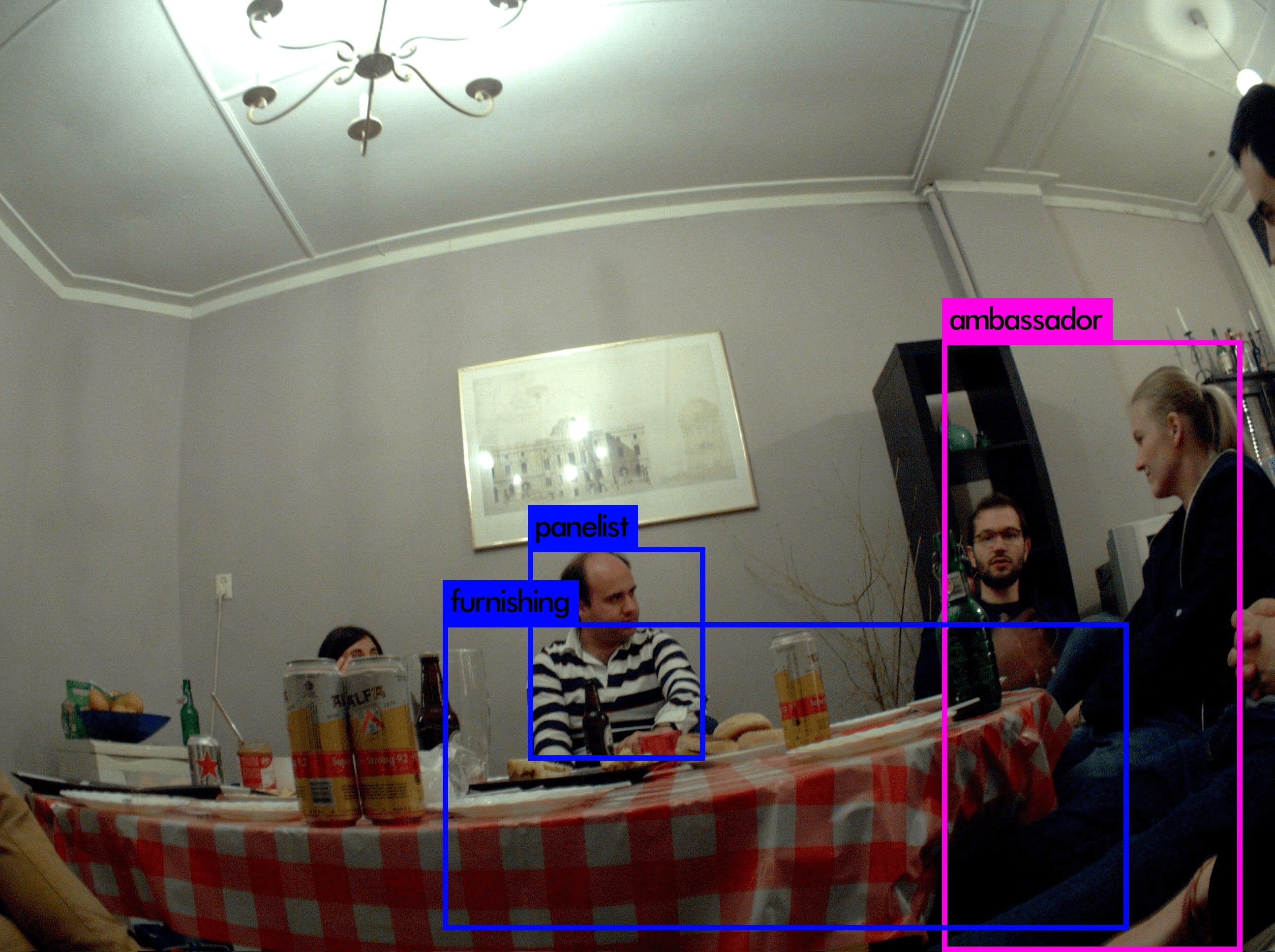}
\begin{tabular}{|p{1.9cm}|p{0.9cm}|p{3.95cm}|p{0.8cm}|}
\hline
\textbf{YOLO9000} & \textbf{Model} & \textbf{Caption} & \textbf{SF} \\ \hline
\multirow{3}{*}{\begin{tabular}[c]{@{}l@{}}{\textit{[}}\textit{Panelist,}\\ \textit{Ambassador,}\\ \textit{Furnishing}{\textit{]}}\end{tabular}} & \textbf{SAT} & \begin{tabular}[c]{@{}l@{}}\textit{A man is standing in front}\\ \textit{of a television}.\end{tabular} & \textbf{0.31} \\ \cline{2-4} 
 & \textbf{NOC} & \begin{tabular}[c]{@{}l@{}}\textit{A man in a kitchen with a}\\ \textit{large mirror.}\end{tabular} & 0.22 \\ \cline{2-4} 
 & \textbf{DNOC} & \begin{tabular}[c]{@{}l@{}}\textit{A man in a kitchen with a}\\ \textit{bottle}.\end{tabular} & 0.19 \\ \hline
\end{tabular}
\end{table}

\vspace{-2em}

\begin{table}[H]
\textbf{2)} \adjustimage{width=3cm,valign=c}{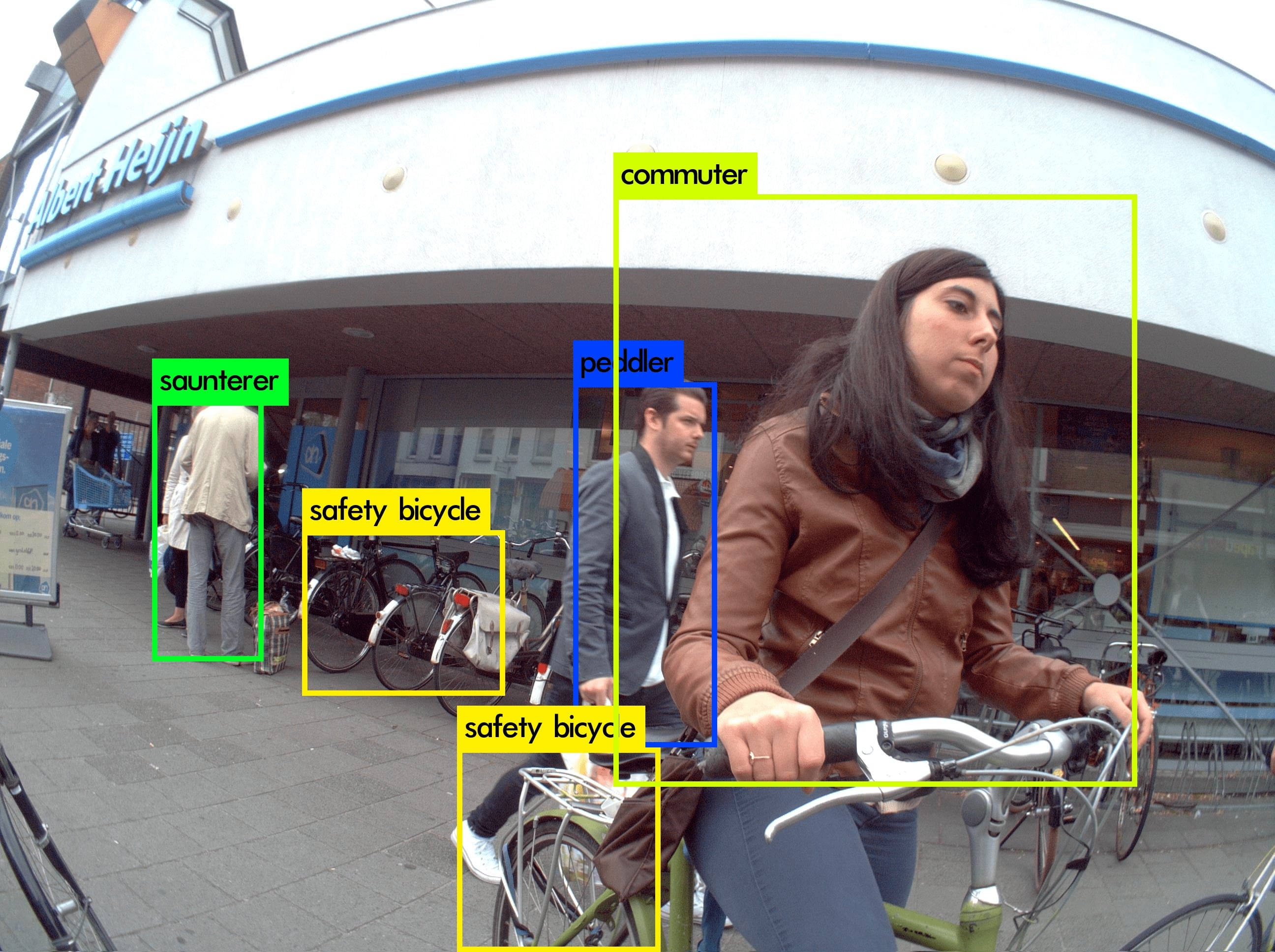}
\begin{tabular}{|p{1.9cm}|p{0.9cm}|p{3.95cm}|p{0.8cm}|}
\hline
\textbf{YOLO9000} & \textbf{Model} & \textbf{Caption} & \textbf{SF} \\ \hline
\multirow{3}{*}{\begin{tabular}[c]{@{}l@{}}{\textit{[}}\textit{Saunterer,}  \\ \textit{Peddler,} \\ \textit{Safety bicycle,} \\ \textit{Commuter}{\textit{]}}\end{tabular}} & \textbf{SAT} & \begin{tabular}[c]{@{}l@{}}\textit{A man and a woman are}\\ \textit{riding a bike.}\end{tabular} & 0.36 \\ \cline{2-4} 
 & \textbf{NOC} & \begin{tabular}[c]{@{}l@{}}\textit{A man is standing on a}\\ \textit{skateboard in the middle}\\ \textit{of a street.}\end{tabular} & \textbf{0.46} \\ \cline{2-4} 
 & \textbf{DNOC} & \begin{tabular}[c]{@{}l@{}}\textit{A man and woman}\\ \textit{sitting on a bicycle.}\end{tabular} & 0.38 \\ \hline
\end{tabular}
\end{table}

\vspace{-2em}
\begin{table}[H]
\textbf{3)} \adjustimage{width=3cm,valign=c}{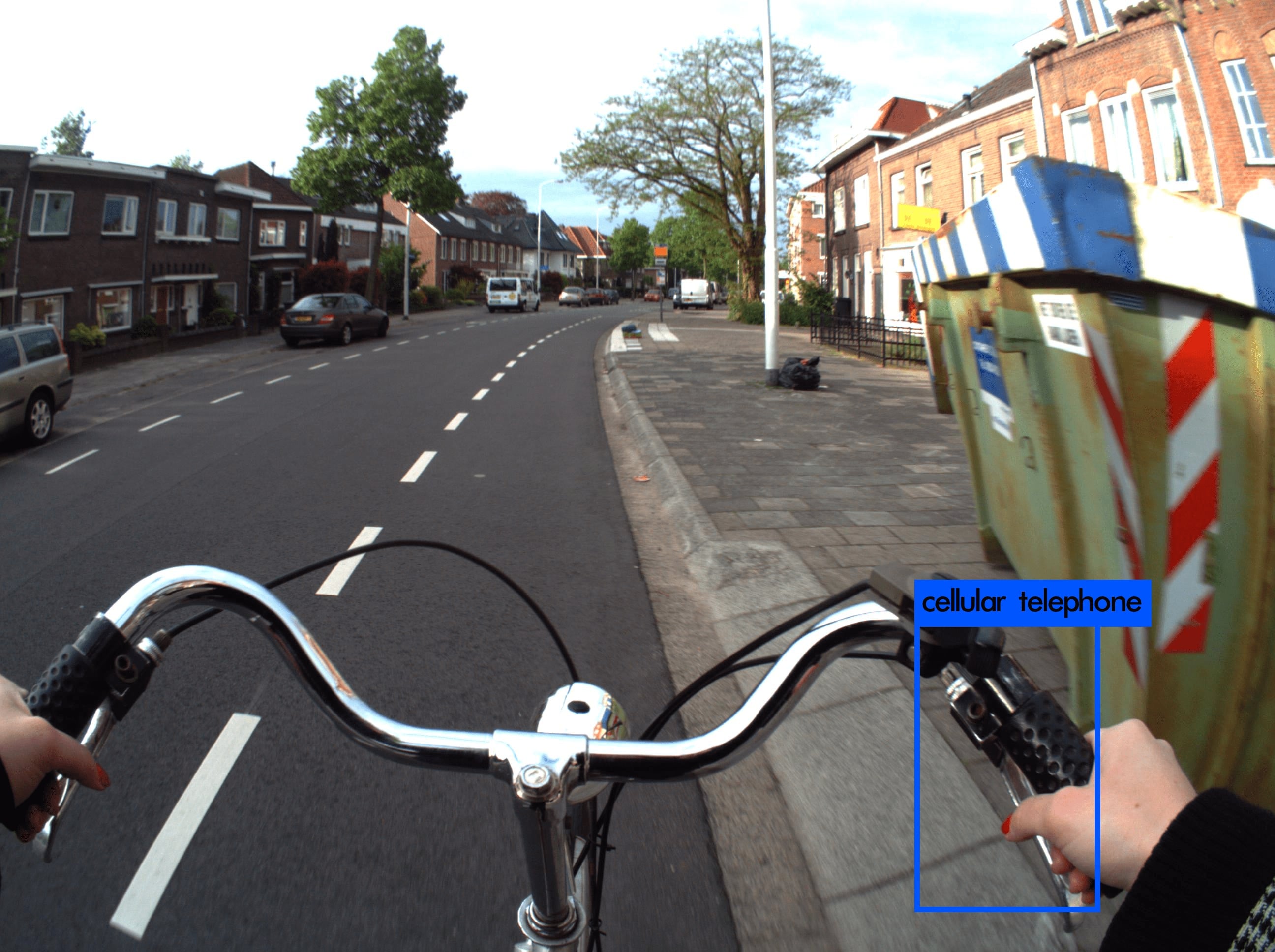}
\begin{tabular}{|p{1.9cm}|p{0.9cm}|p{3.95cm}|p{0.8cm}|}
\hline
\textbf{YOLO9000} & \textbf{Model} & \textbf{Caption} & \textbf{SF} \\ \hline
\multirow{3}{*}{\begin{tabular}[c]{@{}l@{}}{\textit{[}}\textit{Cellular} \\ \textit{Telephone}{\textit{]}}\end{tabular}} & \textbf{SAT} & \begin{tabular}[c]{@{}l@{}}\textit{A person riding a skateboard}\\ \textit{down a street.}\end{tabular} & \textbf{0.24} \\ \cline{2-4} 
 & \textbf{NOC} & \begin{tabular}[c]{@{}l@{}}\textit{A woman is sitting on a}\\ \textit{bench with her bike.}\end{tabular} & 0.18 \\ \cline{2-4} 
 & \textbf{DNOC} & \begin{tabular}[c]{@{}l@{}}\textit{A bicycle parked on a}\\ \textit{sidewalk near a  street.}\end{tabular} & 0.21 \\ \hline
\end{tabular}
\end{table}

\vspace{-2em}
\begin{table}[H]
\textbf{4)} \adjustimage{width=3cm,valign=c}{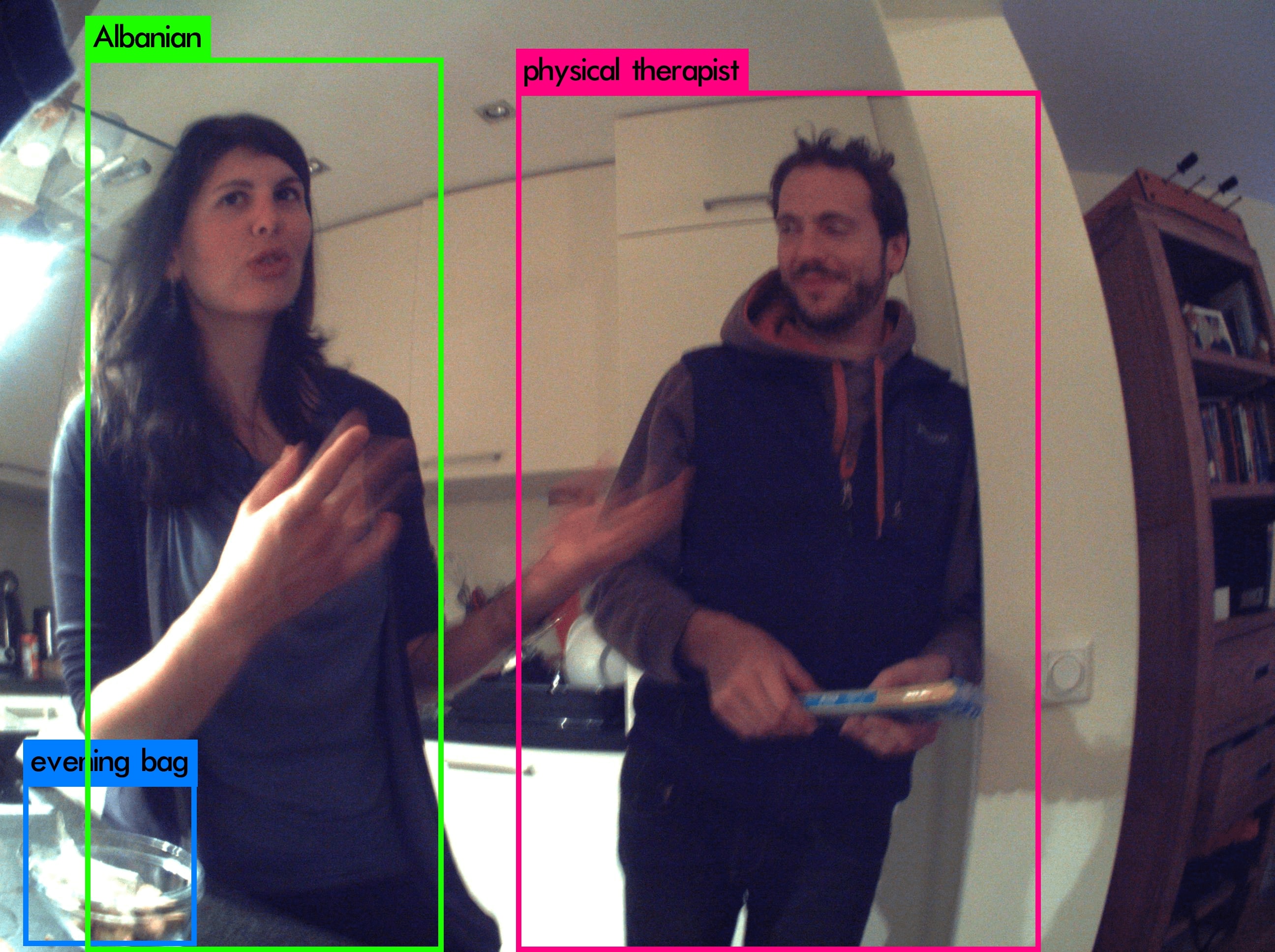}
\begin{tabular}{|p{1.9cm}|p{0.9cm}|p{3.95cm}|p{0.8cm}|}
\hline
\textbf{YOLO9000} & \textbf{Model} & \textbf{Caption} & \textbf{SF} \\ \hline
\multirow{3}{*}{\begin{tabular}[c]{@{}l@{}}{\textit{[}}\textit{Evening} \textit{bag},\\ \textit{Albanian}, \\ \textit{Physical} \\ \textit{therapist}{\textit{]}}\end{tabular}} & \textbf{SAT} & \begin{tabular}[c]{@{}l@{}}\textit{A woman holding a nintendo}\\ \textit{wii game controller.}\end{tabular} & 0.44 \\ \cline{2-4} 
 & \textbf{NOC} & \begin{tabular}[c]{@{}l@{}}\textit{A woman standing in a}\\ \textit{bathroom holding a wii}\\ \textit{remote.}\end{tabular} & 0.53 \\ \cline{2-4} 
 & \textbf{DNOC} & \begin{tabular}[c]{@{}l@{}}\textit{A man in a black shirt and}\\ \textit{a cell phone.}\end{tabular} & \textbf{0.54} \\ \hline
\end{tabular}
\end{table}

\vspace{-2em}
\begin{table}[H]
\textbf{5)} \adjustimage{width=3cm,valign=c}{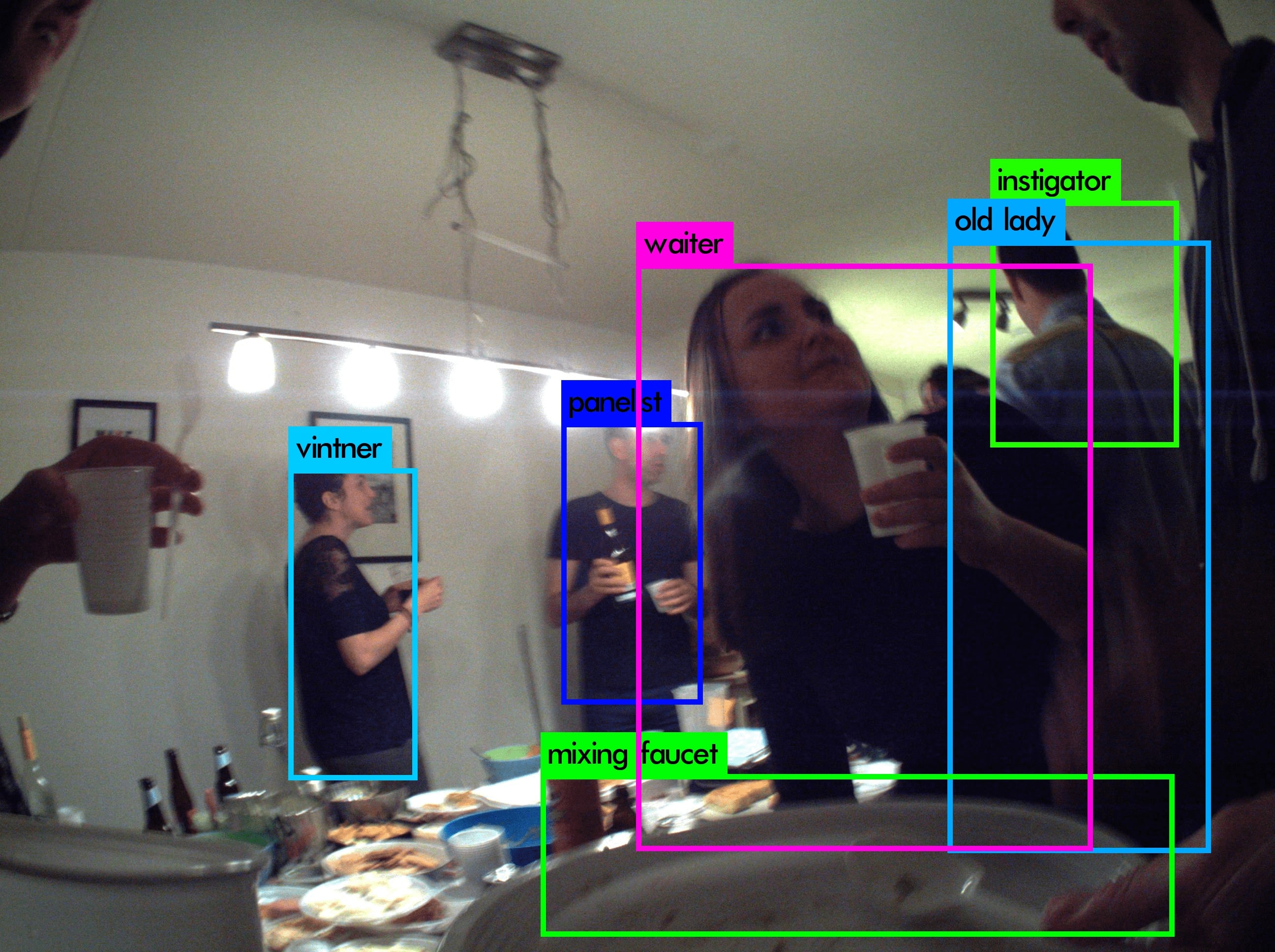}
\begin{tabular}{|p{1.9cm}|p{0.9cm}|p{3.95cm}|p{0.8cm}|}
\hline
\textbf{YOLO9000} & \textbf{Model} & \textbf{Caption} & \textbf{SF} \\ \hline
\multirow{3}{*}{\begin{tabular}[c]{@{}l@{}}{\textit{[}}\textit{Instigator},\\ \textit{Panelist},\\ \textit{Old lady}, \\ \textit{Vintner}, \\ \textit{Mixing faucet},\\ \textit{Waiter}{\textit{]}}\end{tabular}} & \textbf{SAT} & \begin{tabular}[c]{@{}l@{}}\textit{A group of people}\\ \textit{playing a video game.}\end{tabular} & 0.14 \\ \cline{2-4} 
 & \textbf{NOC} & \begin{tabular}[c]{@{}l@{}}\textit{A man in a kitchen}\\ \textit{preparing food in a}\\ \textit{kitchen}.\end{tabular} & \textbf{0.39} \\ \cline{2-4} 
 & \textbf{DNOC} & \begin{tabular}[c]{@{}l@{}}\textit{A group of people}\\ \textit{standing around a table.}\end{tabular} & 0.13 \\ \hline
\end{tabular}
\end{table}

\vspace{-2em}
\begin{table}[H]
\textbf{6)} \adjustimage{width=3cm,valign=c}{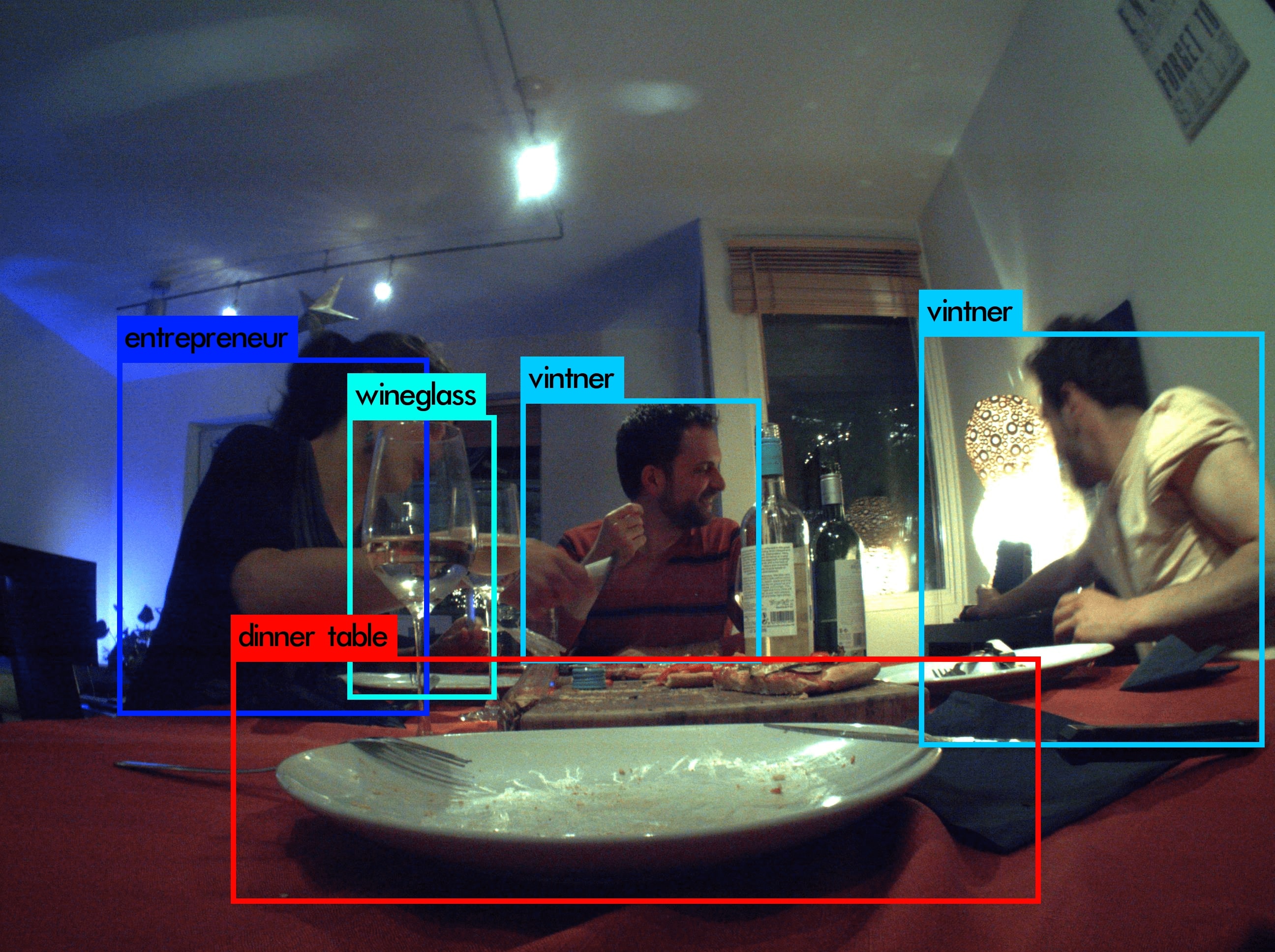}
\begin{tabular}{|p{1.9cm}|p{0.9cm}|p{3.95cm}|p{0.8cm}|}
\hline
\textbf{YOLO9000} & \textbf{Model} & \textbf{Caption} & \textbf{SF} \\ \hline
\multirow{3}{*}{\begin{tabular}[c]{@{}l@{}}{\textit{[}}\textit{Entrepreneur}, \\ \textit{Wineglass},\\ \textit{Vintner},\\  \textit{Dinner table}{\textit{]}}\end{tabular}} & \textbf{SAT} & \begin{tabular}[c]{@{}l@{}}\textit{A group of people sitting}\\ \textit{at a table with wine glasses.}\end{tabular} & 0.36 \\ \cline{2-4} 
 & \textbf{NOC} & \begin{tabular}[c]{@{}l@{}}\textit{A group of people sitting}\\ \textit{at a table with food.}\end{tabular} & 0.27 \\ \cline{2-4} 
 & \textbf{DNOC} & \begin{tabular}[c]{@{}l@{}}\textit{A man and woman sitting}\\ \textit{at a table with food.}\end{tabular} & \textbf{0.38} \\ \hline
\end{tabular}
\end{table}

\vspace{-2em}
\begin{table}[H]
\textbf{7)} \adjustimage{width=3cm,valign=c}{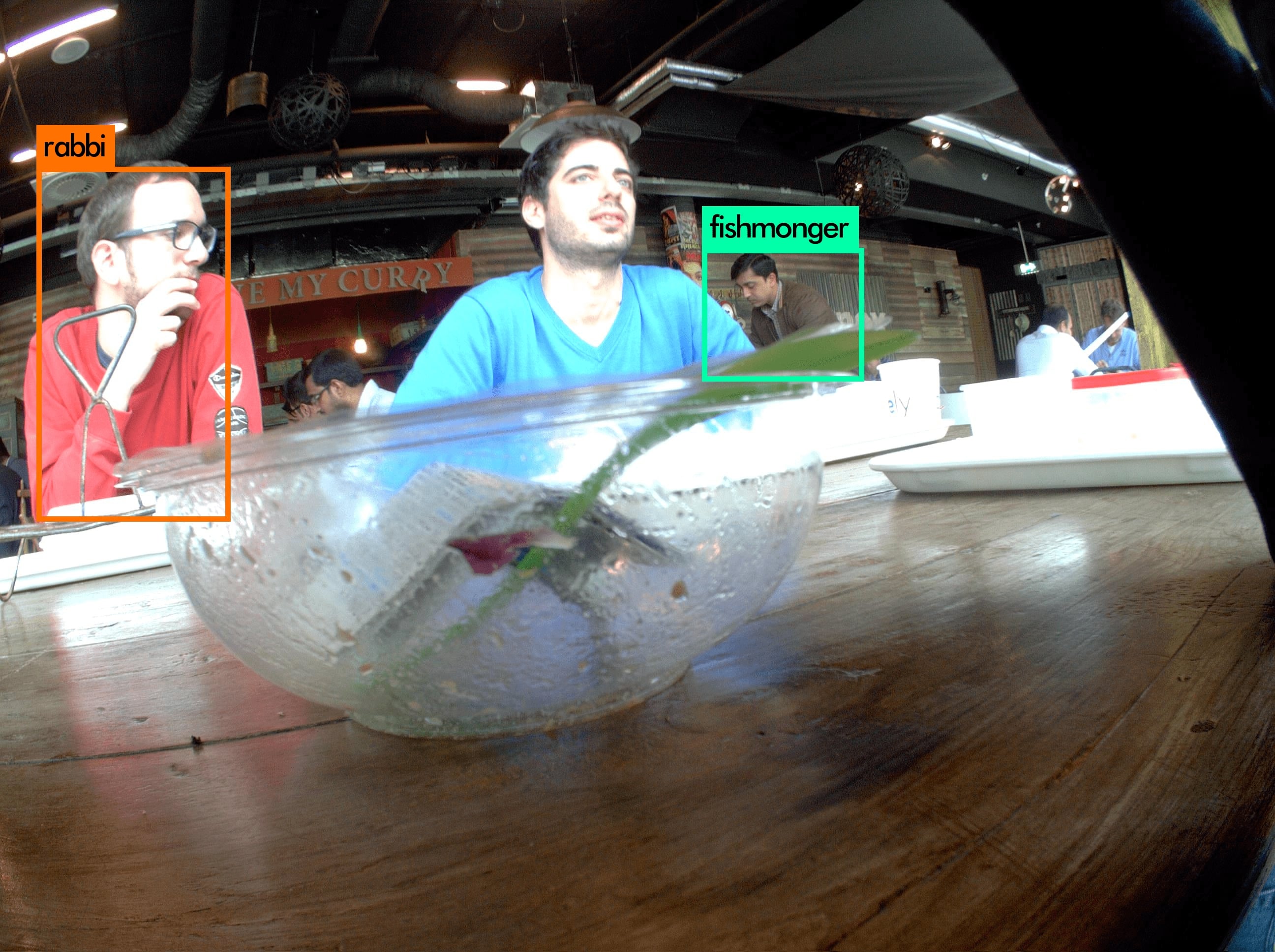}
\begin{tabular}{|p{1.9cm}|p{0.9cm}|p{3.95cm}|p{0.8cm}|}
\hline
\textbf{YOLO9000} & \textbf{Model} & \textbf{Caption} & \textbf{SF} \\ \hline
\multirow{3}{*}{\begin{tabular}[c]{@{}l@{}}{\textit{[}}\textit{Rabbi},\\ \textit{Fishmonger}{\textit{]}}\end{tabular}} & \textbf{SAT} & \begin{tabular}[c]{@{}l@{}}\textit{A woman sitting at a}\\ \textit{table with a glass of wine.}\end{tabular} & 0.18 \\ \cline{2-4} 
 & \textbf{NOC} & \begin{tabular}[c]{@{}l@{}}\textit{A man is standing }\\ \textit{in the middle of a table}\\ \textit{with a bowl of food.}.\end{tabular} & \textbf{0.19} \\ \cline{2-4} 
 & \textbf{DNOC} & \begin{tabular}[c]{@{}l@{}}\textit{A man in a white }\\ \textit{shirt is holding a bowl.}\end{tabular} & 0.18 \\ \hline
\end{tabular}
\end{table}

\vspace{-2em}
\begin{table}[H]
\textbf{8)} \adjustimage{width=3cm,valign=c}{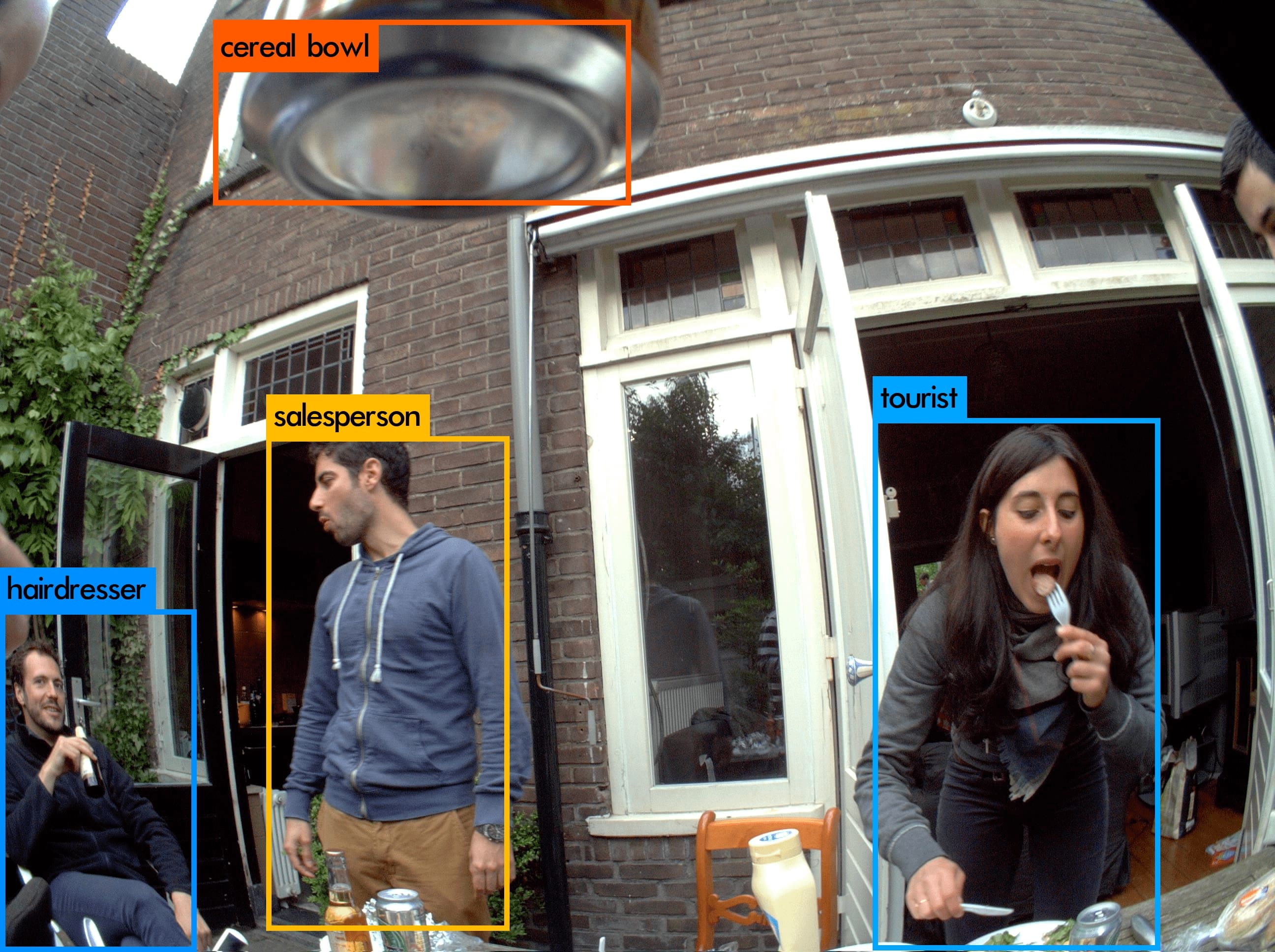}
\begin{tabular}{|p{1.9cm}|p{0.9cm}|p{3.95cm}|p{0.8cm}|}
\hline
\textbf{YOLO9000} & \textbf{Model} & \textbf{Caption} & \textbf{SF} \\ \hline
\multirow{3}{*}{\begin{tabular}[c]{@{}l@{}}{\textit{[}}\textit{Cereal Bowl},\\ \textit{Salesperson},\\ \textit{Tourist}, \\ \textit{Hairdresser}{\textit{]}}\end{tabular}} & \textbf{SAT} & \begin{tabular}[c]{@{}l@{}}\textit{a couple of people }\\ \textit{standing around a table.}\end{tabular} & 0.23 \\ \cline{2-4} 
 & \textbf{NOC} & \begin{tabular}[c]{@{}l@{}}\textit{A man standing next }\\ \textit{to a woman in a city.}\end{tabular} & \textbf{0.33} \\ \cline{2-4} 
 & \textbf{DNOC} & \begin{tabular}[c]{@{}l@{}}\textit{a group of people standing }\\ \textit{around a table.}\end{tabular} & 0.16 \\ \hline
\end{tabular}
\end{table}

\vspace{-2em}
\begin{table}[H]
\textbf{9)} \adjustimage{width=3cm,valign=c}{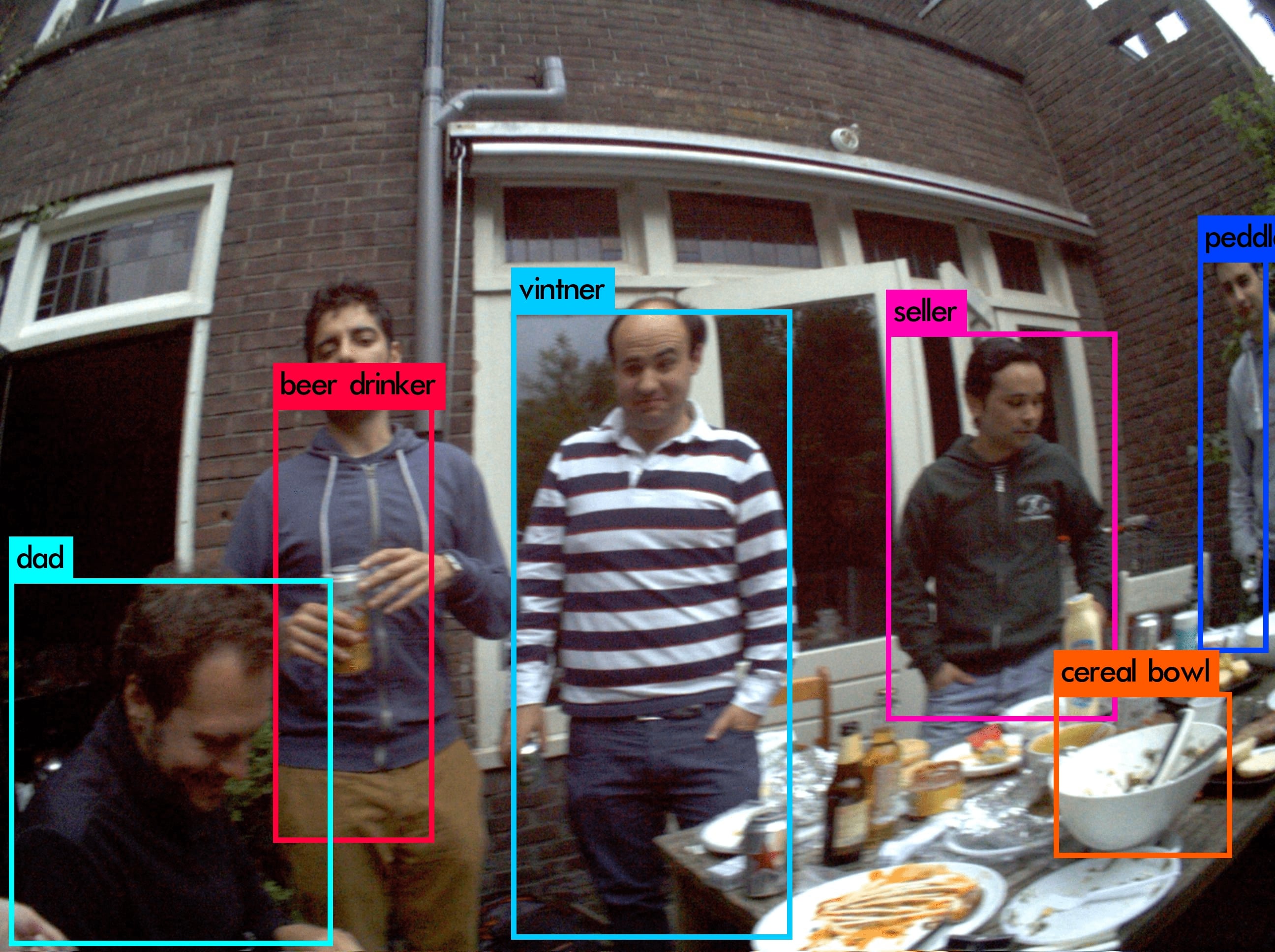}
\begin{tabular}{|p{1.9cm}|p{0.9cm}|p{3.95cm}|p{0.8cm}|}
\hline
\textbf{YOLO9000} & \textbf{Model} & \textbf{Caption} & \textbf{SF} \\ \hline
\multirow{3}{*}{\begin{tabular}[c]{@{}l@{}}{\textit{[}}\textit{Peddler},\\ \textit{Seller},\\ \textit{Beer Drinker}, \\ \textit{Vintner}, \\ \textit{Dad},\\ \textit{Cereal Bowl}{\textit{]}}\end{tabular}} & \textbf{SAT} & \begin{tabular}[c]{@{}l@{}}\textit{a group of people }\\ \textit{standing around a table.}\end{tabular} & 0.1 \\ \cline{2-4} 
 & \textbf{NOC} & \begin{tabular}[c]{@{}l@{}}\textit{A group of people }\\ \textit{standing around a table with }\\ \textit{a large white plate of food.}.\end{tabular} & \textbf{0.34} \\ \cline{2-4} 
 & \textbf{DNOC} & \begin{tabular}[c]{@{}l@{}}\textit{a group of people sitting }\\ \textit{around a table with food.}\end{tabular} & 0.23 \\ \hline
\end{tabular}
\end{table}

\vspace{-2em}
\begin{table}[H]
\textbf{10)} \adjustimage{width=3cm,valign=c}{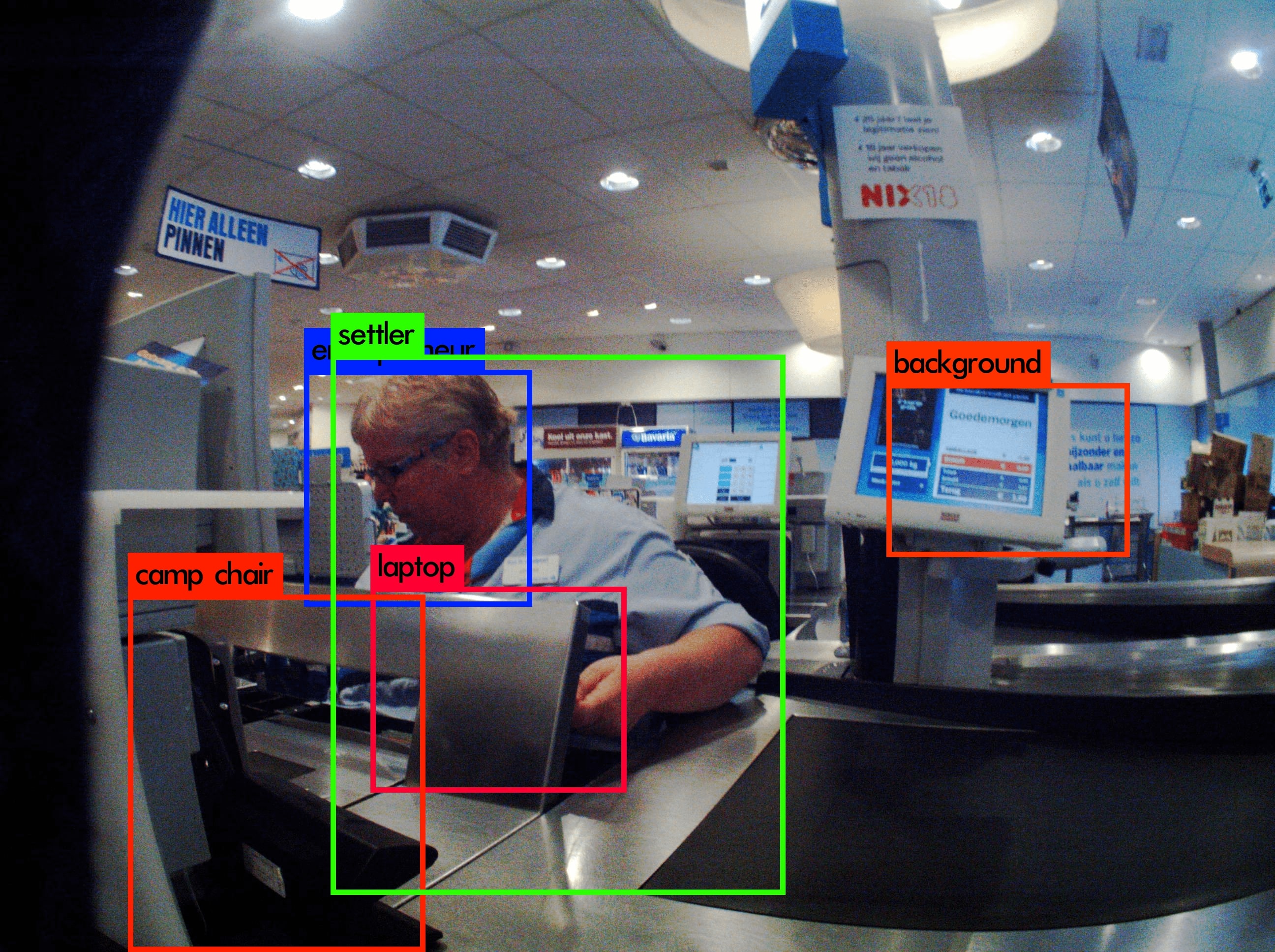}
\begin{tabular}{|p{1.9cm}|p{0.9cm}|p{3.95cm}|p{0.8cm}|}
\hline
\textbf{YOLO9000} & \textbf{Model} & \textbf{Caption} & \textbf{SF} \\ \hline
\multirow{3}{*}{\begin{tabular}[c]{@{}l@{}}{\textit{[}}\textit{Entrepreneur},\\ \textit{Background},\\ \textit{Laptop},\\ \textit{Camp Chair},\\ \textit{Settler}{\textit{]}}\end{tabular}} & \textbf{SAT} & \begin{tabular}[c]{@{}l@{}}\textit{A man sitting at }\\ \textit{a table with a laptop.}\end{tabular} & 0.42 \\ \cline{2-4} 
 & \textbf{NOC} & \begin{tabular}[c]{@{}l@{}}\textit{A man in a kitchen }\\ \textit{with a large display }\\ \textit{of food.}.\end{tabular} & 0.44 \\ \cline{2-4} 
 & \textbf{DNOC} & \begin{tabular}[c]{@{}l@{}}\textit{A man in a suit and }\\ \textit{tv standing in front of a tv.}\end{tabular} & \textbf{0.62} \\ \hline
\end{tabular}
\caption{Captions generated by each of the pre-trained models for images from the Egoshots dataset. The SF metric is used in order to evaluate the captions by taking into consideration the objects detected by YOLO9000. As observed in image 3, due to a poor object detector failing to detect all objects correctly, SF penalize the caption only for its incorrect cosine similarity and skips object diversity (SF=$s_i$, Assumption 1 broken).} 

\label{label1}
\end{table}

\subsection{Examples of Captions generated for uncommonly composed artistic images}

Non real images consisting of uncommon scenarios, such as having numerous objects in unrealistic settings are hard to find in existing public datasets. However, these are excellent samples to challenge current image captioning models and highlight some flaws, such as non generalization to unseen configuration scenes. 

Results below show the limitations of state of the art image captioning and object detector models, where even in the presence of known objects, the object detector fails. It is important to note how in cases where the object detector fails, the metric is not reliable. Since SF would be unable to penalize the captioning model, as it cannot rely on a faithful (i.e. robust enough) object detector (\#O=0, Assumption 2 broken), SF cannot be applied.

\begin{table}[H]
\textbf{1)} \adjustimage{width=3cm,valign=c}{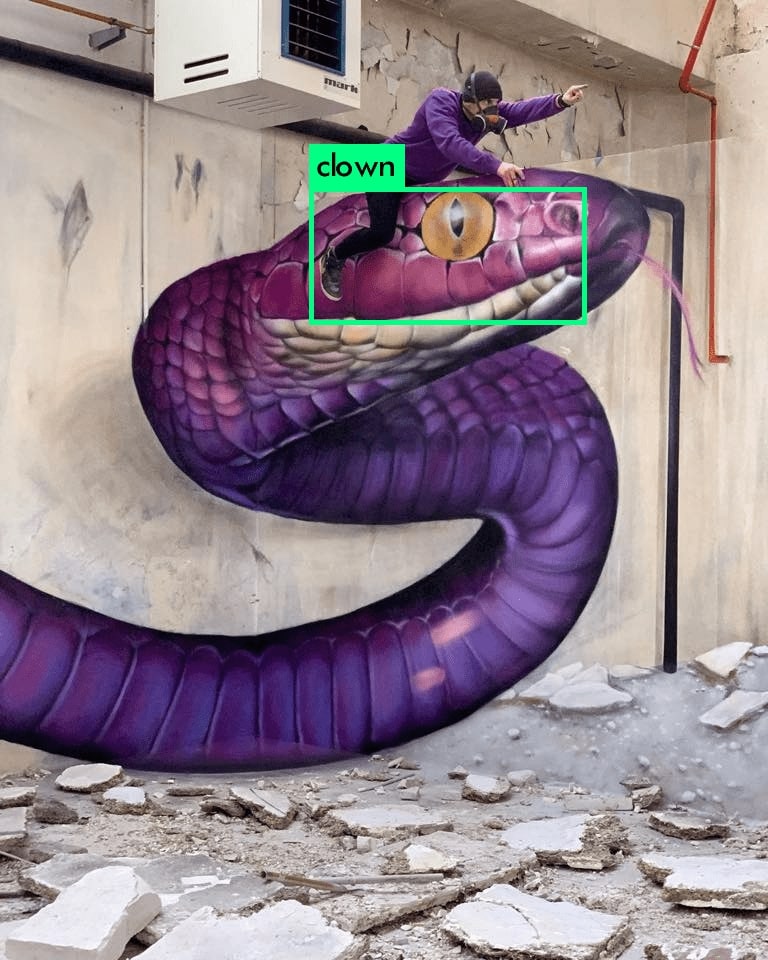}
\begin{tabular}{|p{1.9cm}|p{0.9cm}|p{3.95cm}|p{0.8cm}|}
\hline
\textbf{YOLO9000} & \textbf{Model} & \textbf{Caption} & \textbf{SF} \\ \hline
\multirow{3}{*}{\begin{tabular}[c]{@{}l@{}}{\textit{[}}\textit{Clown}{\textit{]}}\end{tabular}} & \textbf{SAT} & \begin{tabular}[c]{@{}l@{}}\textit{A person wearing a hat and}\\ \textit{and a hat.}.\end{tabular} & 0.45 \\ \cline{2-4} 
 & \textbf{NOC} & \begin{tabular}[c]{@{}l@{}}\textit{A man is sitting on a bed}\\ \textit{with a cat.}\end{tabular} & 0.46 \\ \cline{2-4} 
 & \textbf{DNOC} & \begin{tabular}[c]{@{}l@{}}\textit{A man is holding a bicycle }\\ \textit{in his hand.}\end{tabular} & 0.36 \\ \hline
\end{tabular}
\end{table}

\vspace{-2em}

\begin{table}[H]
\textbf{2)} \adjustimage{width=3cm,valign=c}{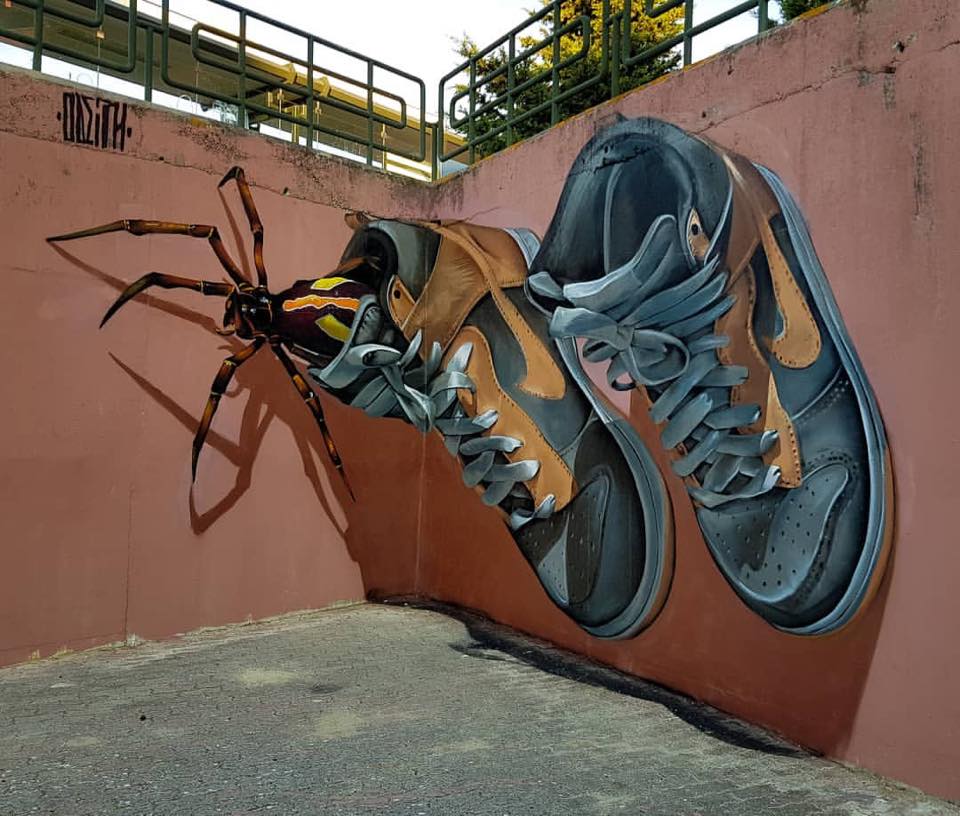}
\begin{tabular}{|p{1.9cm}|p{0.9cm}|p{3.95cm}|p{0.8cm}|}
\hline
\textbf{YOLO9000} & \textbf{Model} & \textbf{Caption} & \textbf{SF} \\ \hline
\multirow{3}{*}{\begin{tabular}[c]{@{}l@{}}{\textit{[}}{\textit{]}}\end{tabular}} & \textbf{SAT} & \begin{tabular}[c]{@{}l@{}}\textit{A person riding a bike}\\ \textit{ on a street.}\end{tabular} & -- \\ \cline{2-4} 
 & \textbf{NOC} & \begin{tabular}[c]{@{}l@{}}\textit{A man is sitting on a }\\ \textit{motorcycle with a dog of a}\\ \textit{street.}\end{tabular} & -- \\ \cline{2-4} 
 & \textbf{DNOC} & \begin{tabular}[c]{@{}l@{}}\textit{A man is sitting on a }\\ \textit{motorcycle with a dog.}\end{tabular} & -- \\ \hline
\end{tabular}
\end{table}

\begin{table}[H]
\textbf{3)} \adjustimage{width=3cm,valign=c}{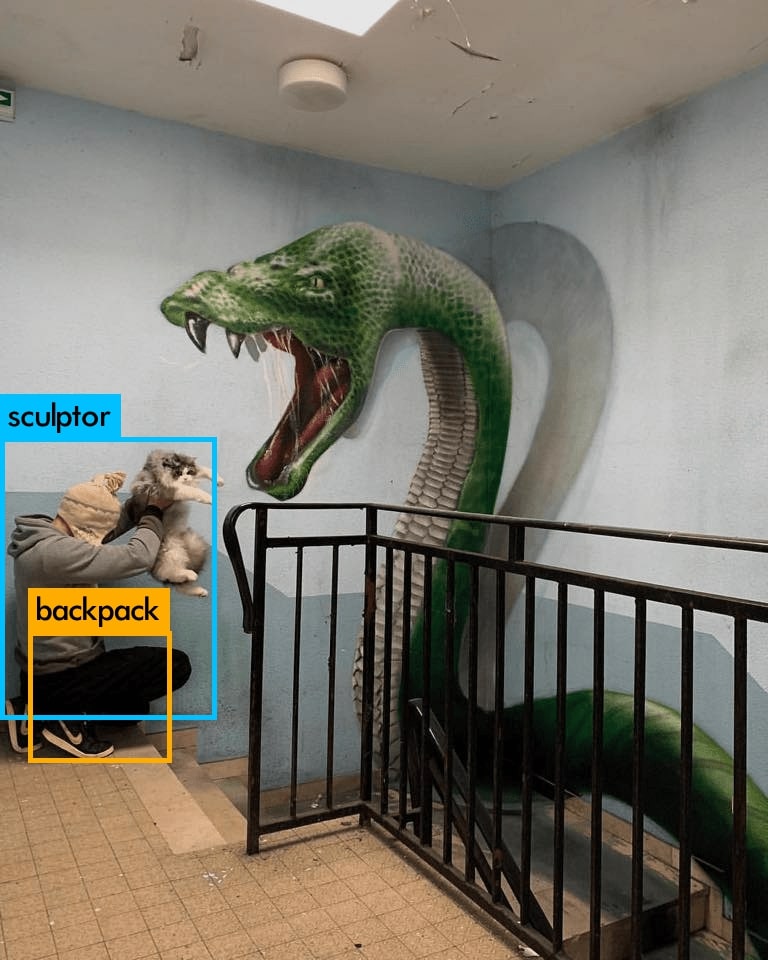}
\begin{tabular}{|p{1.9cm}|p{0.9cm}|p{3.95cm}|p{0.8cm}|}
\hline
\textbf{YOLO9000} & \textbf{Model} & \textbf{Caption} & \textbf{SF} \\ \hline
\multirow{3}{*}{\begin{tabular}[c]{@{}l@{}}{\textit{[}}\textit{Sculptor,}  \\ \textit{Backpack}{\textit{]}}\end{tabular}} & \textbf{SAT} & \begin{tabular}[c]{@{}l@{}}\textit{A young man is doing a trick }\\ \textit{on a skateboard.}\end{tabular} & 0.44 \\ \cline{2-4} 
 & \textbf{NOC} & \begin{tabular}[c]{@{}l@{}}\textit{A man is sitting on a wooden }\\ \textit{bench in a park.}\end{tabular} & 0.39 \\ \cline{2-4} 
 & \textbf{DNOC} & \begin{tabular}[c]{@{}l@{}}\textit{A bird sitting on a bird in a }\\ \textit{room.}\end{tabular} & 0.31 \\ \hline
\end{tabular}
\end{table}

\begin{table}[H]
\textbf{4)} \adjustimage{width=3cm,valign=c}{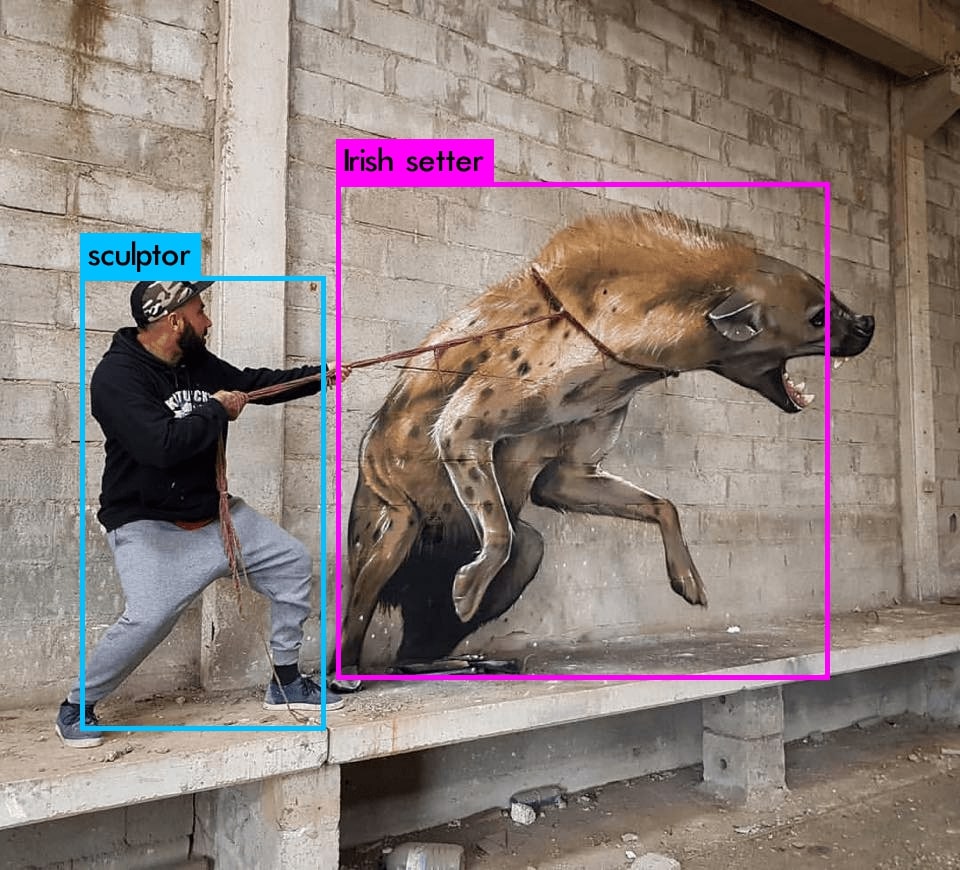}
\begin{tabular}{|p{1.9cm}|p{0.9cm}|p{3.95cm}|p{0.8cm}|}
\hline
\textbf{YOLO9000} & \textbf{Model} & \textbf{Caption} & \textbf{SF} \\ \hline
\multirow{3}{*}{\begin{tabular}[c]{@{}l@{}}{\textit{[}}\textit{Irish setter,}  \\ \textit{Sculptor}{\textit{]}}\end{tabular}} & \textbf{SAT} & \begin{tabular}[c]{@{}l@{}}\textit{A man sitting on a bench }\\ \textit{with a dog.}\end{tabular} & 0.37 \\ \cline{2-4} 
 & \textbf{NOC} & \begin{tabular}[c]{@{}l@{}}\textit{A man is standing on a }\\ \textit{skateboard in the air.}\end{tabular} & 0.25 \\ \cline{2-4} 
 & \textbf{DNOC} & \begin{tabular}[c]{@{}l@{}}\textit{A man is riding a dog in a }\\ \textit{zoo.}\end{tabular} & 0.33 \\ \hline
\end{tabular}
\end{table}

\begin{table}[H]
\textbf{5)} \adjustimage{width=3cm,valign=c}{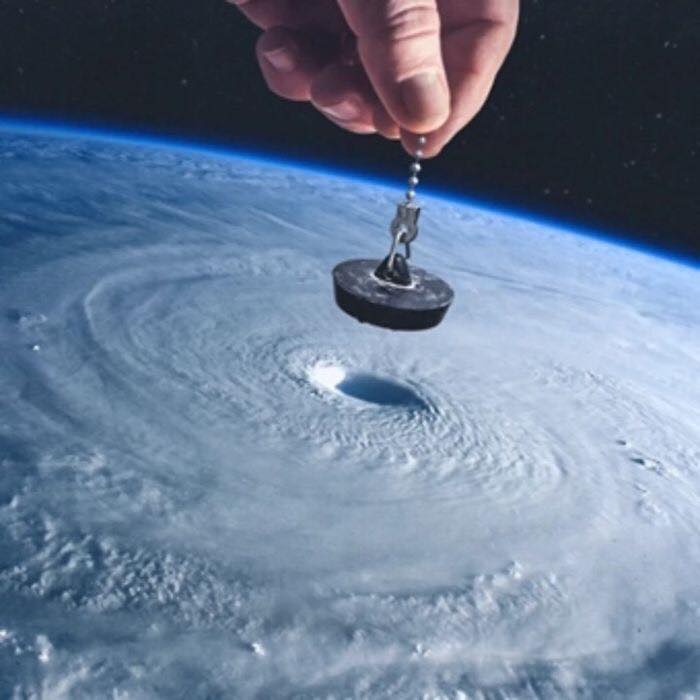}
\begin{tabular}{|p{1.9cm}|p{0.9cm}|p{3.95cm}|p{0.8cm}|}
\hline
\textbf{YOLO9000} & \textbf{Model} & \textbf{Caption} & \textbf{SF} \\ \hline
\multirow{3}{*}{\begin{tabular}[c]{@{}l@{}}{\textit{[}}{\textit{]}}\end{tabular}} & \textbf{SAT} & \begin{tabular}[c]{@{}l@{}}\textit{A person on a skateboard in }\\ \textit{the air.}\end{tabular} & -- \\ \cline{2-4} 
 & \textbf{NOC} & \begin{tabular}[c]{@{}l@{}}\textit{A man is sitting on a white }\\ \textit{and blue shirt.} \end{tabular} & -- \\ \cline{2-4} 
 & \textbf{DNOC} & \begin{tabular}[c]{@{}l@{}}\textit{A person is sitting on a }\\ \textit{bicycle in the snow.}\end{tabular} & -- \\ \hline
\end{tabular}
\end{table}

\begin{table}[H]
\textbf{6)} \adjustimage{width=3cm,valign=c}{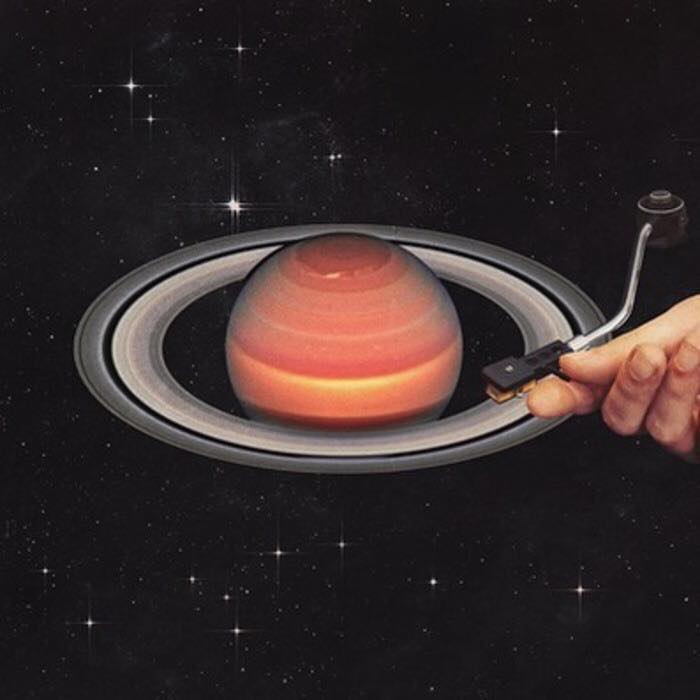}
\begin{tabular}{|p{1.9cm}|p{0.9cm}|p{3.95cm}|p{0.8cm}|}
\hline
\textbf{YOLO9000} & \textbf{Model} & \textbf{Caption} & \textbf{SF} \\ \hline
\multirow{3}{*}{\begin{tabular}[c]{@{}l@{}}{\textit{[}}{\textit{]}}\end{tabular}} & \textbf{SAT} & \begin{tabular}[c]{@{}l@{}}\textit{A close up of a person }\\ \textit{holding a banana.}\end{tabular} & -- \\ \cline{2-4} 
 & \textbf{NOC} & \begin{tabular}[c]{@{}l@{}}\textit{A man is sitting on a pink }\\ \textit{metal plate.}\end{tabular} & -- \\ \cline{2-4} 
 & \textbf{DNOC} & \begin{tabular}[c]{@{}l@{}}\textit{A bowl with a bowl in it is }\\ \textit{sitting on a table.}\end{tabular} & -- \\ \hline
\end{tabular}
\end{table}

\begin{table}[H]
\textbf{7)} \adjustimage{width=3cm,valign=c}{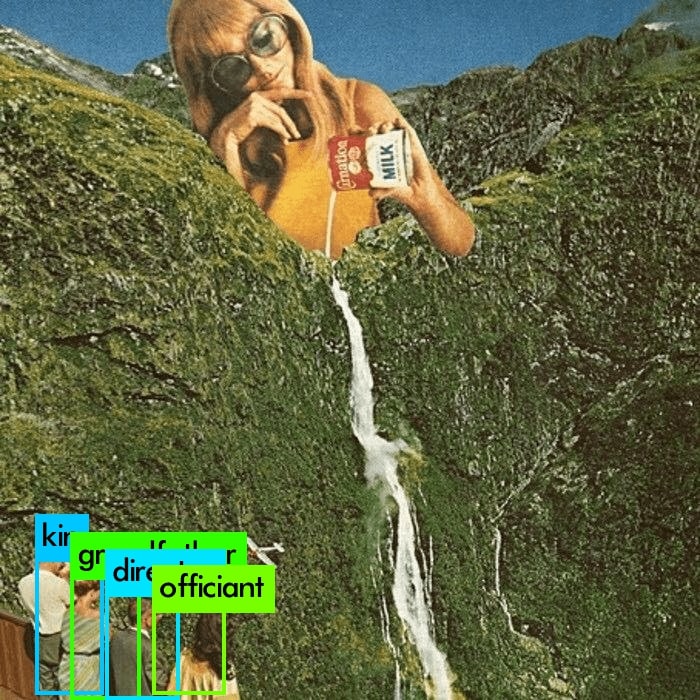}
\begin{tabular}{|p{1.9cm}|p{0.9cm}|p{3.95cm}|p{0.8cm}|}
\hline
\textbf{YOLO9000} & \textbf{Model} & \textbf{Caption} & \textbf{SF} \\ \hline
\multirow{3}{*}{\begin{tabular}[c]{@{}l@{}}{\textit{[}}\textit{Kin,}  \\ \textit{Grandfather,} \\ \textit{Director,} \\ \textit{Officiant}  {\textit{]}}\end{tabular}} & \textbf{SAT} & \begin{tabular}[c]{@{}l@{}}\textit{A close up of a person }\\ \textit{holding a frisbee.}\end{tabular} & 0.15 \\ \cline{2-4} 
 & \textbf{NOC} & \begin{tabular}[c]{@{}l@{}}\textit{A man is riding a skateboard }\\ \textit{on a hill.}\end{tabular} & 0.26 \\ \cline{2-4} 
 & \textbf{DNOC} & \begin{tabular}[c]{@{}l@{}}\textit{A woman is holding a bicyle }\\ \textit{in the woods.}\end{tabular} & 0.15 \\ \hline
\end{tabular}
\end{table}

\begin{table}[H]
\textbf{8)} \adjustimage{width=3cm,valign=c}{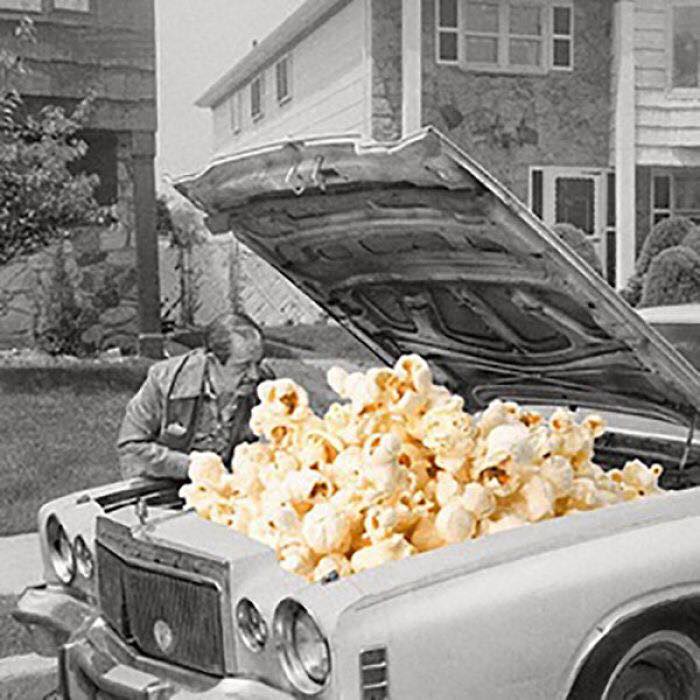}
\begin{tabular}{|p{1.9cm}|p{0.9cm}|p{3.95cm}|p{0.8cm}|}
\hline
\textbf{YOLO9000} & \textbf{Model} & \textbf{Caption} & \textbf{SF} \\ \hline
\multirow{3}{*}{\begin{tabular}[c]{@{}l@{}}{\textit{[}}{\textit{]}}\end{tabular}} & \textbf{SAT} & \begin{tabular}[c]{@{}l@{}}\textit{A person is holding a piece }\\ \textit{of food.}\end{tabular} & -- \\ \cline{2-4} 
 & \textbf{NOC} & \begin{tabular}[c]{@{}l@{}}\textit{A man is sitting on a white}\\ \textit{and black dog.}\end{tabular} & -- \\ \cline{2-4} 
 & \textbf{DNOC} & \begin{tabular}[c]{@{}l@{}}\textit{A car with a car on it and a }\\ \textit{car.}\end{tabular} & -- \\ \hline
\end{tabular}
\end{table}

\begin{table}[H]
\textbf{9)} \adjustimage{width=3cm,valign=c}{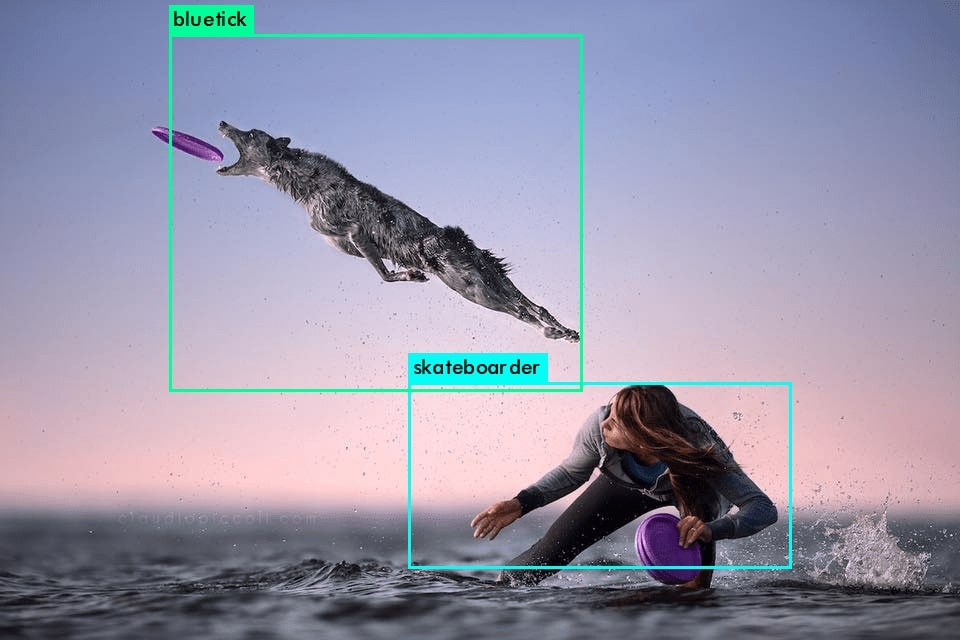}
\begin{tabular}{|p{1.9cm}|p{0.9cm}|p{3.95cm}|p{0.8cm}|}
\hline
\textbf{YOLO9000} & \textbf{Model} & \textbf{Caption} & \textbf{SF} \\ \hline
\multirow{3}{*}{\begin{tabular}[c]{@{}l@{}}{\textit{[}}\textit{Skateboarder,}  \\ \textit{Bluetick}{\textit{]}}\end{tabular}} & \textbf{SAT} & \begin{tabular}[c]{@{}l@{}}\textit{A person on a snowboard}\\ \textit{in the snow.}\end{tabular} & 0.32 \\ \cline{2-4} 
 & \textbf{NOC} & \begin{tabular}[c]{@{}l@{}}\textit{A man is riding a surfboard}\\ \textit{on a wave.}\end{tabular} & 0.34 \\ \cline{2-4} 
 & \textbf{DNOC} & \begin{tabular}[c]{@{}l@{}}\textit{A man riding a wave on top}\\ \textit{of a frisbee.}\end{tabular} & 0.26 \\ \hline
\end{tabular}
\end{table}

\begin{table}[H]
\textbf{10)} \adjustimage{width=3cm,valign=c}{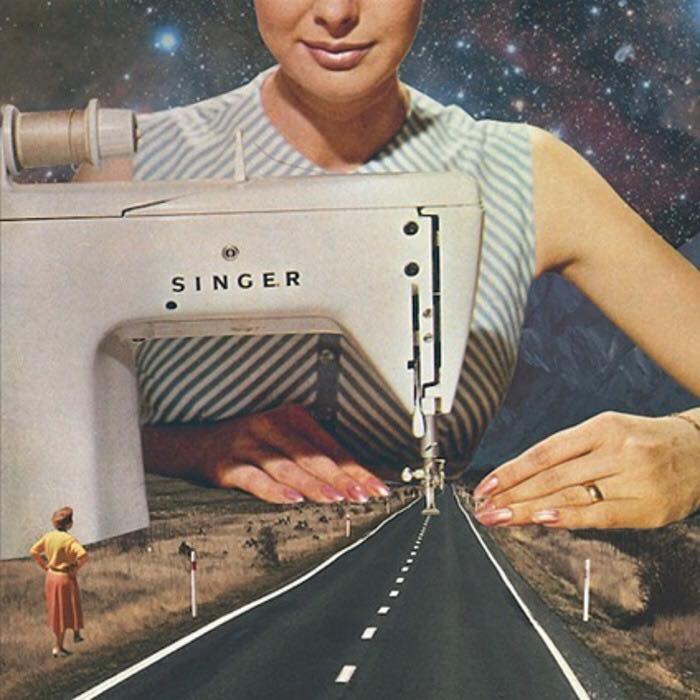}
\begin{tabular}{|p{1.9cm}|p{0.9cm}|p{3.95cm}|p{0.8cm}|}
\hline
\textbf{YOLO9000} & \textbf{Model} & \textbf{Caption} & \textbf{SF} \\ \hline
\multirow{3}{*}{\begin{tabular}[c]{@{}l@{}}{\textit{[}}{\textit{]}}\end{tabular}} & \textbf{SAT} & \begin{tabular}[c]{@{}l@{}}\textit{A man standing in front of a }\\ \textit{microwave oven.}\end{tabular} & -- \\ \cline{2-4} 
 & \textbf{NOC} & \begin{tabular}[c]{@{}l@{}}\textit{A man is sitting on a table }\\ \textit{with a large cake.}\end{tabular} & -- \\ \cline{2-4} 
 & \textbf{DNOC} & \begin{tabular}[c]{@{}l@{}}\textit{A man is holding a bicycle in }\\ \textit{his hand.}\end{tabular} & -- \\ \hline
\end{tabular}

\caption{Showcasing the failure of pre-trained neural networks on non real images. 
}
\end{table}

\subsection{Metric Limitations} 
We must note some limitations of the metric, which should be complemented/extended to (1) account for verbs and other syntactic elements of the caption; (2) rate a caption in terms of the quality of the interpretation, taking into account the count of objects of the same type in the image with respect to those present in the caption. Particular models for counting such as \citep{Cohen17} is a particular example on how to enhance the label-less dataset annotation pipeline proposed here.


Metrics should be evaluated in more targeted application use cases, e.g. the usefulness of such captions for targeted users such as the blind, in concrete applications such as navigation settings \citep{Weiss19}.

\subsection{Acknowledgements}
We thank Valerio Iebba, Gemma Roig and David Filliat for practical and insightful feedback. We thank the site \textit{QuickTurtles}\footnote{\url{www.fb.com/QuickTurtles}} for sharing fun images worth feeding to a neural network to stress-test them.

\end{document}

%% file: math_commands.tex
\usepackage{amsmath,amsfonts,bm}

\def\eqref#1{equation~\ref{#1}}

\def\1{\bm{1}}

\DeclareMathAlphabet{\mathsfit}{\encodingdefault}{\sfdefault}{m}{sl}
\SetMathAlphabet{\mathsfit}{bold}{\encodingdefault}{\sfdefault}{bx}{n}













%% file: ICLR-ML-IRL.bbl
\begin{thebibliography}{44}
\providecommand{\natexlab}[1]{#1}
\providecommand{\url}[1]{\texttt{#1}}
\expandafter\ifx\csname urlstyle\endcsname\relax
  \providecommand{\doi}[1]{doi: #1}\else
  \providecommand{\doi}{doi: \begingroup \urlstyle{rm}\Url}\fi

\bibitem[Agrawal et~al.(2019)Agrawal, Desai, Wang, Jain, Chen, Batra, Johnson,
  Parikh, Anderson, and Lee]{Agrawal2019nocapsNO}
Harsh Agrawal, Karan Desai, Yufei Wang, Rishabh Jain, Xinlei Chen, Dhruv Batra,
  Mark~S Johnson, Devi Parikh, Peter Anderson, and Stefan Lee.
\newblock {nocaps: novel object captioning at scale}.
\newblock \emph{International Conference on Computer Vision}, abs/1812.08658,
  2019.

\bibitem[Anderson et~al.(2017)Anderson, He, Buehler, Teney, Johnson, Gould, and
  Zhang]{Anderson2017BottomUpAT}
Peter Anderson, Xiaodong He, Chris Buehler, Damien Teney, Mark Johnson, Stephen
  Gould, and Lei Zhang.
\newblock {Bottom-Up and Top-Down Attention for Image Captioning and Visual
  Question Answering}.
\newblock \emph{2018 IEEE/CVF Conference on Computer Vision and Pattern
  Recognition}, pp.\  6077--6086, 2017.

\bibitem[Banerjee \& Lavie(2005)Banerjee and Lavie]{Banerjee2005METEORAA}
Satanjeev Banerjee and Alon Lavie.
\newblock {METEOR: An Automatic Metric for {MT} Evaluation with Improved
  Correlation with Human Judgments}.
\newblock In \emph{IEEvaluation@ACL}, 2005.

\bibitem[Conneau et~al.(2017)Conneau, Lample, Ranzato, Denoyer, and
  J{\'e}gou]{Conneau2017WordTW}
Alexis Conneau, Guillaume Lample, Marc'Aurelio Ranzato, Ludovic Denoyer, and
  Herv{\'e} J{\'e}gou.
\newblock {Word Translation Without Parallel Data}.
\newblock \emph{ArXiv}, abs/1710.04087, 2017.

\bibitem[Demirel et~al.(2019)Demirel, Cinbis, and
  Ikizler-Cinbis]{Demirel2019ImageCW}
Berkan Demirel, Ramazan~Gokberk Cinbis, and Nazli Ikizler-Cinbis.
\newblock {Image Captioning with Unseen Objects}.
\newblock \emph{ArXiv}, abs/1908.00047, 2019.

\bibitem[Donahue et~al.(2015)Donahue, Hendricks, Guadarrama, Rohrbach,
  Venugopalan, Darrell, and Saenko]{Donahue2015LongtermRC}
Jeff Donahue, Lisa~Anne Hendricks, Sergio Guadarrama, Marcus Rohrbach,
  Subhashini Venugopalan, Trevor Darrell, and Kate Saenko.
\newblock {Long-term recurrent convolutional networks for visual recognition
  and description}.
\newblock In \emph{CVPR}, 2015.

\bibitem[Duan et~al.(2019)Duan, Bai, Xie, Qi, Huang, and
  Tian]{Duan2019CenterNetKT}
Kaiwen Duan, Song Bai, Lingxi Xie, Honggang Qi, Qingming Huang, and Qi~Tian.
\newblock {CenterNet: Keypoint Triplets for Object Detection}.
\newblock \emph{ArXiv}, abs/1904.08189, 2019.

\bibitem[Everingham et~al.()Everingham, Van~Gool, Williams, Winn, and
  Zisserman]{pascal-voc-2012}
M.~Everingham, L.~Van~Gool, C.~K.~I. Williams, J.~Winn, and A.~Zisserman.
\newblock The {PASCAL} {V}isual {O}bject {C}lasses {C}hallenge 2012 {(VOC2012)}
  {R}esults.
\newblock URL
  \url{http://www.pascal-network.org/challenges/VOC/voc2012/workshop/index.html}.

\bibitem[Fan \& Crandall(2016)Fan and Crandall]{Fan2016DeepDiaryAC}
Chenyou Fan and David~J. Crandall.
\newblock {DeepDiary: Automatic Caption Generation for Lifelogging Image
  Streams}.
\newblock \emph{ArXiv}, abs/1608.03819, 2016.

\bibitem[Feng et~al.(2018)Feng, Ma, Liu, and Luo]{Feng2018UnsupervisedIC}
Yang Feng, Lin Ma, Wei Liu, and Jiebo Luo.
\newblock {Unsupervised Image Captioning}.
\newblock In \emph{CVPR}, 2018.

\bibitem[Gurari et~al.(2018)Gurari, Li, Stangl, Guo, Lin, Grauman, Luo, and
  Bigham]{Gurari2018VizWizGC}
Danna Gurari, Qing Li, Abigale~J. Stangl, Anhong Guo, Chi Lin, Kristen Grauman,
  Jiebo Luo, and Jeffrey~P. Bigham.
\newblock {VizWiz Grand Challenge: Answering Visual Questions from Blind
  People}.
\newblock \emph{2018 IEEE/CVF Conference on Computer Vision and Pattern
  Recognition}, pp.\  3608--3617, 2018.
\newblock URL \url{https://arxiv.org/abs/1802.08218}.

\bibitem[Karpathy \& Feifei(2014)Karpathy and Feifei]{Karpathy2014DeepVA}
Andrej Karpathy and Li~Feifei.
\newblock {Deep Visual-Semantic Alignments for Generating Image Descriptions}.
\newblock \emph{2015 IEEE Conference on Computer Vision and Pattern Recognition
  (CVPR)}, pp.\  3128--3137, 2014.

\bibitem[Kenigsfield \& El-Yaniv(2019)Kenigsfield and
  El-Yaniv]{Kenigsfield2019LeveragingAT}
Gal~Sadeh Kenigsfield and Ran El-Yaniv.
\newblock {Leveraging Auxiliary Text for Deep Recognition of Unseen Visual
  Relationships}.
\newblock \emph{ArXiv}, abs/1910.12324, 2019.

\bibitem[Kuznetsova et~al.(2018)Kuznetsova, Rom, Alldrin, Uijlings, Krasin,
  Pont-Tuset, Kamali, Popov, Malloci, Duerig, and Ferrari]{Kuznetsova2018TheOI}
Alina Kuznetsova, Hassan Rom, Neil Alldrin, Jasper R.~R. Uijlings, Ivan Krasin,
  Jordi Pont-Tuset, Shahab Kamali, Stefan Popov, Matteo Malloci, Tom Duerig,
  and Vittorio Ferrari.
\newblock {The Open Images Dataset V4: Unified image classification, object
  detection, and visual relationship detection at scale}.
\newblock \emph{ArXiv}, abs/1811.00982, 2018.
\newblock URL \url{https://arxiv.org/abs/1811.00982}.

\bibitem[Li et~al.(2017)Li, Tang, Deng, Zhang, and Tian]{Li2017ImageCW}
Linghui Li, Sheng Tang, Lixi Deng, Yongdong Zhang, and Qi~Tian.
\newblock {Image Caption with Global-Local Attention}.
\newblock In \emph{AAAI}, 2017.

\bibitem[Liang et~al.(2020)Liang, Guan, and Rojas]{Liang2020VisualSemanticGA}
Zhijun Liang, Yi-Sheng Guan, and Juan Rojas.
\newblock {Visual-Semantic Graph Attention Network for Human-Object Interaction
  Detection}.
\newblock 2020.

\bibitem[Lin(2004)]{Lin2004ROUGEAP}
Chin-Yew Lin.
\newblock {ROUGE: A Package For Automatic Evaluation Of Summaries}.
\newblock In \emph{ACL 2004}, 2004.

\bibitem[Lin et~al.(2014)Lin, Maire, Belongie, Bourdev, Girshick, Hays, Perona,
  Ramanan, Zitnick, and Doll{\'a}r]{Lin2014MicrosoftCC}
Tsung-Yi Lin, Michael Maire, Serge~J. Belongie, Lubomir~D. Bourdev, Ross~B.
  Girshick, James Hays, Pietro Perona, Deva Ramanan, C.~Lawrence Zitnick, and
  Piotr Doll{\'a}r.
\newblock {Microsoft COCO: Common Objects in Context}.
\newblock In \emph{ECCV}, 2014.
\newblock URL \url{http://arxiv.org/abs/1405.0312}.

\bibitem[Liu et~al.(2016)Liu, Anguelov, Erhan, Szegedy, Reed, Fu, and
  Berg]{Liu2016SSDSS}
Wei Liu, Dragomir Anguelov, Dumitru Erhan, Christian Szegedy, Scott~E. Reed,
  Cheng-Yang Fu, and Alexander~C. Berg.
\newblock {SSD: Single Shot MultiBox Detector}.
\newblock In \emph{ECCV}, 2016.

\bibitem[Lu et~al.(2016)Lu, Xiong, Parikh, and Socher]{Lu2016KnowingWT}
Jiasen Lu, Caiming Xiong, Devi Parikh, and Richard Socher.
\newblock {Knowing When to Look: Adaptive Attention via a Visual Sentinel for
  Image Captioning}.
\newblock \emph{2017 IEEE Conference on Computer Vision and Pattern Recognition
  (CVPR)}, pp.\  3242--3250, 2016.

\bibitem[Lu et~al.(2018)Lu, Yang, Batra, and Parikh]{Lu2018NeuralBT}
Jiasen Lu, Jianwei Yang, Dhruv Batra, and Devi Parikh.
\newblock {Neural Baby Talk}.
\newblock \emph{2018 IEEE/CVF Conference on Computer Vision and Pattern
  Recognition}, pp.\  7219--7228, 2018.

\bibitem[Mikolov et~al.(2013)Mikolov, Sutskever, Chen, Corrado, and
  Dean]{Mikolov2013DistributedRO}
Tomas Mikolov, Ilya Sutskever, Kai Chen, Gregory~S. Corrado, and Jeffrey Dean.
\newblock {Distributed Representations of Words and Phrases and their
  Compositionality}.
\newblock In \emph{NIPS}, 2013.

\bibitem[Nagarajan et~al.(2020)Nagarajan, Li, Feichtenhofer, and
  Grauman]{Nagarajan2020EGOTOPOEA}
Tushar Nagarajan, Yanghao Li, Christoph Feichtenhofer, and Kristen Grauman.
\newblock {EGO-TOPO: Environment Affordances from Egocentric Video}.
\newblock \emph{ArXiv}, abs/2001.04583, 2020.

\bibitem[Olivastri et~al.(2019)Olivastri, Singh, and
  Cuzzolin]{Olivastri2019EndtoEndVC}
Silvio Olivastri, Gurkirt Singh, and Fabio Cuzzolin.
\newblock {End-to-End Video Captioning}.
\newblock 2019.

\bibitem[Papineni et~al.(2001)Papineni, Roukos, Ward, and
  Zhu]{Papineni2001BleuAM}
Kishore Papineni, Salim Roukos, Todd Ward, and Wei-Jing Zhu.
\newblock {Bleu: a Method for Automatic Evaluation of Machine Translation}.
\newblock In \emph{ACL}, 2001.

\bibitem[Paul~Cohen et~al.(2017)Paul~Cohen, Boucher, Glastonbury, Lo, and
  Bengio]{Cohen17}
Joseph Paul~Cohen, Genevieve Boucher, Craig~A. Glastonbury, Henry~Z. Lo, and
  Yoshua Bengio.
\newblock {Count-ception: Counting by Fully Convolutional Redundant Counting}.
\newblock In \emph{The IEEE International Conference on Computer Vision (ICCV)
  Workshops}, Oct 2017.

\bibitem[Redmon \& Farhadi(2016)Redmon and Farhadi]{Redmon2016YOLO9000BF}
Joseph Redmon and Ali Farhadi.
\newblock {YOLO9000: Better, Faster, Stronger}.
\newblock \emph{2017 IEEE Conference on Computer Vision and Pattern Recognition
  (CVPR)}, pp.\  6517--6525, 2016.

\bibitem[Redmon \& Farhadi(2018)Redmon and Farhadi]{Redmon2018YOLOv3AI}
Joseph Redmon and Ali Farhadi.
\newblock {YOLOv3: An Incremental Improvement}.
\newblock \emph{ArXiv}, abs/1804.02767, 2018.

\bibitem[Ren et~al.(2017)Ren, Wang, Zhang, Lv, and Li]{Ren2017DeepRL}
Zhou Ren, Xiaoyu Wang, Ning Zhang, Xutao Lv, and Li-Jia Li.
\newblock {Deep Reinforcement Learning-Based Image Captioning with Embedding
  Reward}.
\newblock \emph{2017 IEEE Conference on Computer Vision and Pattern Recognition
  (CVPR)}, pp.\  1151--1159, 2017.

\bibitem[Rohrbach et~al.(2018)Rohrbach, Hendricks, Burns, Darrell, and
  Saenko]{Rohrbach18}
Anna Rohrbach, Lisa~Anne Hendricks, Kaylee Burns, Trevor Darrell, and Kate
  Saenko.
\newblock {Object Hallucination in Image Captioning}.
\newblock \emph{CoRR}, abs/1809.02156, 2018.
\newblock URL \url{http://arxiv.org/abs/1809.02156}.

\bibitem[Selvaraju et~al.(2019)Selvaraju, Lee, Shen, Jin, Batra, and
  Parikh]{Selvaraju2019TakingAH}
Ramprasaath~R. Selvaraju, Stefan Lee, Yilin Shen, Hongxia Jin, Dhruv Batra, and
  Devi Parikh.
\newblock {Taking a HINT: Leveraging Explanations to Make Vision and Language
  Models More Grounded}.
\newblock \emph{2019 IEEE/CVF International Conference on Computer Vision
  (ICCV)}, pp.\  2591--2600, 2019.

\bibitem[Tan et~al.(2018)Tan, Sun, Kong, Zhang, Yang, and Liu]{Tan2018ASO}
Chuanqi Tan, Fuchun Sun, Tao Kong, Wenchang Zhang, Chao Yang, and Chunfang Liu.
\newblock {A Survey on Deep Transfer Learning}.
\newblock \emph{ArXiv}, abs/1808.01974, 2018.

\bibitem[Tsutsui et~al.(2019)Tsutsui, Zhi, Reza, Crandall, and
  Yu]{Tsutsui2019ActiveOM}
Satoshi Tsutsui, Dian Zhi, Md~Alimoor Reza, David~J. Crandall, and Chen Yu.
\newblock {Active Object Manipulation Facilitates Visual Object Learning: An
  Egocentric Vision Study}.
\newblock \emph{ArXiv}, abs/1906.01415, 2019.

\bibitem[Vedantam et~al.(2014)Vedantam, Zitnick, and
  Parikh]{Vedantam2014CIDErCI}
Ramakrishna Vedantam, C.~Lawrence Zitnick, and Devi Parikh.
\newblock {CIDEr}: Consensus-based image description evaluation.
\newblock \emph{2015 IEEE Conference on Computer Vision and Pattern Recognition
  (CVPR)}, pp.\  4566--4575, 2014.

\bibitem[Vedantam et~al.(2017)Vedantam, Bengio, Murphy, Parikh, and
  Chechik]{Vedantam2017ContextAwareCF}
Ramakrishna Vedantam, Samy Bengio, Kevin Murphy, Devi Parikh, and Gal Chechik.
\newblock {Context-Aware Captions from Context-Agnostic Supervision}.
\newblock \emph{2017 IEEE Conference on Computer Vision and Pattern Recognition
  (CVPR)}, pp.\  1070--1079, 2017.

\bibitem[Venugopalan et~al.(2016)Venugopalan, Hendricks, Rohrbach, Mooney,
  Darrell, and Saenko]{Venugopalan2016CaptioningIW}
Subhashini Venugopalan, Lisa~Anne Hendricks, Marcus Rohrbach, Raymond~J.
  Mooney, Trevor Darrell, and Kate Saenko.
\newblock {Captioning Images with Diverse Objects}.
\newblock \emph{2017 IEEE Conference on Computer Vision and Pattern Recognition
  (CVPR)}, pp.\  1170--1178, 2016.

\bibitem[Vinyals et~al.(2014)Vinyals, Toshev, Bengio, and
  Erhan]{Vinyals2014ShowAT}
Oriol Vinyals, Alexander Toshev, Samy Bengio, and Dumitru Erhan.
\newblock {Show and tell: A neural image caption generator}.
\newblock \emph{2015 IEEE Conference on Computer Vision and Pattern Recognition
  (CVPR)}, pp.\  3156--3164, 2014.

\bibitem[Wang et~al.(2018)Wang, Madhyastha, and Specia]{Wang2018ObjectCB}
Josiah Wang, Pranava~Swaroop Madhyastha, and Lucia Specia.
\newblock {Object Counts! Bringing Explicit Detections Back into Image
  Captioning}.
\newblock In \emph{NAACL-HLT}, 2018.

\bibitem[Weiss et~al.(2019)Weiss, Chamorro, Girgis, Luck, Kahou, Cohen,
  Nowrouzezahrai, Precup, Golemo, and Pal]{Weiss19}
Martin Weiss, Sim{\'o}n Chamorro, Roger Girgis, Margaux Luck, Samira~Ebrahimi
  Kahou, Joseph~Paul Cohen, Derek Nowrouzezahrai, Doina Precup, Florian Golemo,
  and Chris Pal.
\newblock {Navigation Agents for the Visually Impaired: A Sidewalk Simulator
  and Experiments}.
\newblock \emph{ArXiv}, abs/1910.13249, 2019.

\bibitem[Wu et~al.(2018)Wu, Zhu, Jiang, and Yang]{Wu2018DecoupledNO}
Yuehua Wu, Linchao Zhu, Lu~Jiang, and Yi~Yang.
\newblock {Decoupled Novel Object Captioner}.
\newblock In \emph{ACM Multimedia}, 2018.

\bibitem[Xu et~al.(2015)Xu, Ba, Kiros, Cho, Courville, Salakhutdinov, Zemel,
  and Bengio]{Xu2015ShowAA}
Kelvin Xu, Jimmy Ba, Ryan Kiros, Kyunghyun Cho, Aaron~C. Courville, Ruslan
  Salakhutdinov, Richard~S. Zemel, and Yoshua Bengio.
\newblock {Show, Attend and Tell: Neural Image Caption Generation with Visual
  Attention}.
\newblock In \emph{ICML}, 2015.

\bibitem[You et~al.(2016)You, Jin, Wang, Fang, and Luo]{You2016ImageCW}
Quanzeng You, Hailin Jin, Zhaowen Wang, Chen Fang, and Jiebo Luo.
\newblock {Image Captioning with Semantic Attention}.
\newblock \emph{2016 IEEE Conference on Computer Vision and Pattern Recognition
  (CVPR)}, pp.\  4651--4659, 2016.

\bibitem[Young et~al.(2014)Young, Lai, Hodosh, and
  Hockenmaier]{Young2014FromID}
Peter Young, Alice Lai, Micah Hodosh, and Julia Hockenmaier.
\newblock {From image descriptions to visual denotations: New similarity
  metrics for semantic inference over event descriptions}.
\newblock \emph{Transactions of the Association for Computational Linguistics},
  2:\penalty0 67--78, 2014.

\bibitem[Zhang et~al.(2020)Zhang, Tawari, Martin, and
  Crandall]{Zhang2020InteractionGF}
Zehua Zhang, Ashish Tawari, Sujitha Martin, and David~J. Crandall.
\newblock {Interaction Graphs for Object Importance Estimation in On-road
  Driving Videos}.
\newblock \emph{ArXiv}, abs/2003.06045, 2020.

\end{thebibliography}
